\pdfoutput=1
\documentclass[11pt]{article}

\usepackage[]{acl}
\usepackage{times}
\usepackage{latexsym}

\usepackage[T1]{fontenc}
\usepackage[utf8]{inputenc}

\usepackage{microtype}

\usepackage{booktabs}
\usepackage{amsmath, amsfonts}
\usepackage{graphicx}
\usepackage{multirow}
\usepackage{enumitem}
\usepackage{hyphenat}
\usepackage{todonotes}
\setlist[itemize]{leftmargin=*}
\setlist[enumerate]{leftmargin=*}

\newboolean{bdemo}
\setboolean{bdemo}{false}
\newcommand{\setbdemotrue}[1]{\textbf{#1}}
\newcommand{\setbdemofalse}[1]{#1}
\newcommand{\bdemo}[1]{\ifthenelse{\boolean{bdemo}}{\setbdemotrue{#1}}{\setbdemofalse{#1}}}

\title{{Memorisation versus Generalisation in Pre-trained Language Models}}

\author{Michael T\"{a}nzer \\
  Imperial College London \\
  \texttt{\small m.tanzer@imperial.ac.uk} \\\And
  Sebastian Ruder\thanks{\hspace{0.2cm}Work done prior to joining Google.} \\
  Google Research\\
  \texttt{\small ruder@google.com} \\\And
  Marek Rei \\
  Imperial College London \\
  \texttt{\small marek.rei@imperial.ac.uk} \\}

\date{}

\begin{document}
\maketitle
\begin{abstract}
State-of-the-art pre-trained language models have been shown to memorise facts and perform well with limited amounts of training data. To gain a better understanding of how these models learn, we study their generalisation and memorisation capabilities in noisy and low-resource scenarios. We find that the training of these models is almost unaffected by label noise and that it is possible to reach near-optimal results even on extremely noisy datasets. 
However, our experiments also show that they mainly learn from high-frequency patterns and largely fail when tested on low-resource tasks such as few-shot learning and rare entity recognition. To mitigate such limitations, we propose an extension based on prototypical networks that improves performance in low-resource named entity recognition tasks.
\end{abstract}

\section{Introduction}\label{sec:intro}
With recent advances in pre-trained language models \cite{Peters2018elmo,devlin_bert_2019,liu_roberta_2019,he2020deberta}, the field of natural language processing has seen improvements in a wide range of tasks and applications. Having acquired general-purpose knowledge from large amounts of unlabelled data, such methods have been shown to learn effectively with limited labelled data for downstream tasks \cite{Howard2018ulmfit} and to generalise well to out-of-distribution examples \cite{hendrycks_pretrained_2020}.

Previous work has extensively studied \emph{what} such models learn, e.g. the types of relational or linguistic knowledge \cite{tenney2019bert,jawahar2019does,Rogers2020}. However, the process of \emph{how} these models learn from downstream data and the qualitative nature of their learning dynamics remain unclear.
Better understanding of the learning processes in these widely-used models is needed in order to know in which scenarios they will fail and how to improve them towards more robust language representations.

The fine-tuning process in pre-trained language models such as BERT \cite{devlin_bert_2019} aims to strike a balance between generalisation and memorisation.
For many applications it is important for the model to generalise---to learn the common patterns in the task while discarding irrelevant noise and outliers.
However, rejecting everything that occurs infrequently is not a reliable learning strategy and in many low-resource scenarios memorisation can be crucial to performing well on a task \cite{tu-etal-2020-empirical}.
By constructing experiments that allow for full control over these parameters, we are able to study the learning dynamics of models in conditions of high label noise or low label frequency.
%This allows us to probe and compare the robustness of the model's behaviour at different levels of noise and sparsity. 
To our knowledge, this is the first qualitative study of the learning behaviour of pre-trained transformer-based language models in conditions of extreme label scarcity and label noise.

%As humans are able to solve novel task variations with only a few examples, we should strive for our models to be able to do the same.
%However, in some situations memorization is necessary component of the model. 
%When training the model on low-resource tasks, with only a small number of labeled examples available for a particular class, the model should still learn. 
%In our experiments we demonstrate that BERT is particularly good at this, with generalisation and memorisation happening in separate stages during training.

%Two key events that happen during learning are the \emph{memorisation} of patterns and the possible \emph{forgetting} of already acquired information \cite{zhang_understanding_2017, toneva_empirical_2019}.

We find that models such as BERT are particularly good at learning general-purpose patterns as generalisation and memorisation become separated into distinct phases during their fine-tuning.
We also observe that the main learning phase is followed by a distinct performance plateau for several epochs before the model starts to memorise the noise. 
This makes the models more robust with regard to the number of training epochs and allows for noisy examples in the data to be identified based only on their training loss.
%We also find that useful examples are learned largely throughout the first few epochs while BERT memorises noise mostly in the later stages of training.

However, we find that these excellent generalisation properties come at the cost of poor performance in few-shot scenarios with extreme class imbalances. Our experiments show that BERT is not able to learn from individual examples and may never predict a particular label until the number of training instances passes a critical threshold. For example, on the \texttt{CoNLL03} \cite{sang_introduction_2003} dataset it requires 25 instances of a class to learn to predict it at all and 100 examples to predict it with some accuracy.
To address this limitation, we propose a method based on prototypical networks \cite{snell_prototypical_2017} that augments BERT with a layer that classifies test examples by finding their closest class centroid. 
%The layer explicitly clusters examples on a per-class basis in feature space and classifies test examples by finding their closest class centroid. 
The method considerably outperforms BERT in challenging training conditions with label imbalances, such as the \texttt{WNUT17} \cite{derczynski_results_2017} rare entities dataset.
%, with less than 100 examples of the minority class on the \texttt{CoNLL03} \cite{sang_introduction_2003} dataset, and provides a slight improvement on the full \texttt{CoNLL03} dataset.

Our contributions are the following: 1) We identify a second phase of learning where BERT does not overfit to noisy datasets. 2) We present experimental evidence that BERT is particularly robust to label noise and can reach near-optimal performance even with extremely strong label noise.
3) We study forgetting in BERT and verify that it is dramatically less forgetful than some alternative methods. 4) We empirically observe that BERT completely fails to recognise minority classes when the number of examples is limited and we propose a new model, ProtoBERT, which outperforms BERT on few-shot versions of \texttt{CoNLL03} and \texttt{JNLPBA}, as well as on the \texttt{WNUT17} dataset.

\section{Previous work}\label{sec:prev-work}

% \todo{MR: This section at the moment is a collection of one-liners about different papers, without a clear narrative. Group related papers together into a paragraph, then add an intro sentence to that paragraph that establishes what this paragraph is going to be about and/or specifies how this is relevant to our work.
% I've structured the intro to now be about generalisation and memorisation, so it would be great to have separate related work sections about these topics.}

% \paragraph{Memorisation} 
Several studies have been conducted on neural models' ability to memorise and recall facts seen during their training. \citet{Petroni2019knowledge_bases} showed that pre-trained language models are surprisingly effective at recalling facts while \citet{carlini_secret_2019} demonstrated that LSTM language models are able to consistently memorise single out-of-distribution (OOD) examples during the very first phase of training and that it is possible to retrieve such examples at test time.
\citet{liu2020early} found that regularising early phases of training is crucial to prevent the studied CNN residual models from memorising noisy examples later on. They also propose a regularisation procedure useful in this setting. Similarly, \citet{li2020gradient} analyse how early stopping and gradient descent affect model robustness to label noise. 

\citet{toneva_empirical_2019}, on the other hand, study forgetting in visual models. They find that models consistently forget a significant portion of the training data and that this fraction of forgettable examples is mainly dependent on intrinsic properties of the training data rather than the specific model. In contrast, we show that a pretrained BERT forgets examples at a dramatically lower rate compared to a BiLSTM and a non-pretrained variant.

% \paragraph{Generalisation}
Memorisation is closely related to generalisation: neural networks have been observed to learn simple patterns before noise \cite{arpit_closer_2017} and generalise despite being able to completely memorise random examples \cite{Zhang2017understanding_deep_learning}. \citet{zhang2021understanding} also show that our current understanding of statistical learning theory cannot explain the super-human generalisation performance of large neural models across many areas of study.

% \paragraph{OOD training}
\citet{hendrycks_pretrained_2020} show that pre-trained models generalise better on out-of-distribution data and are better able to detect such data compared to non-pretrained methods but that they still do not cleanly separate in- and out-of-distribution examples.
% \todo{MR: Unclear how this previous sentence relates to our work. Is this about generalisation? If so, then great, but would be good to add more research on generalisation.}
\citet{kumar2020user} find that pre-trained methods such as BERT are sensitive to spelling noise and typos. In contrast to noise in the input, we focus on the models' learning dynamics in the presence of label noise and find that pre-trained methods are remarkably resilient to such cases.

\section{Experimental setting}\label{sec:experimental-setting}
We investigate the performance of pre-trained language models in specific adverse conditions. In order to evaluate generalisation abilities, we first create datasets with varying levels of label noise by randomly permuting some of the labels in the training data. This procedure allows us to pinpoint noisy examples and evaluate the performance on clean and noisy datapoints separately.
Then, in order to investigate memorisation we train the models on datasets that contain only a small number of examples for a particular class. This allows us to evaluate how well the models are able to learn from individual datapoints as opposed to high-frequency patterns.
We make the code for the experiments available online.\footnote{\url{https://github.com/Michael-Tanzer/BERT-mem-lowres}}

\paragraph{Datasets} We focus on the task of named entity recognition (NER) and employ the \texttt{CoNLL03} \cite{sang_introduction_2003}, the \texttt{JNLPBA} \cite{collier_introduction_2004}, and the \texttt{WNUT17} \cite{derczynski_results_2017} datasets.
NER is commonly used for evaluating pre-trained language models on structured prediction and its natural class imbalance is well suited for our probing experiments.
\texttt{CoNLL03} and \texttt{JNLPBA} are standard datasets for NER and Bio-NER respectively. The \texttt{WNUT17} dataset is motivated by the observation that state-of-the-art methods tend to memorise entities during training \cite{augenstein_generalisation_2017}. The dataset focuses on identifying \bdemo{unusual or rare entities} at test time that cannot be simply memorised by the model. We evaluate based on entity-level F$_1$ unless stated otherwise.
    
\paragraph{Language models} 
We use \bdemo{BERT-base} \cite{devlin_bert_2019} as the main language model for our experiments, as BERT is widely used in practice and other variations of pre-trained language models build on a similar architecture.
The model is augmented with a classification feed-forward layer and fine-tuned using the cross-entropy loss with a learning rate of $10^{-4}$.
AdamW \cite{loshchilov_decoupled_2019} is used during training with weight decay of 0.01 and a linear warm-up rate of 10\%. The test results are recorded using the model that produced the highest validation metrics.
% \todo{MR: This section originally said that the model is trained for 4 epochs. That seems impossible, as the graphs show way more than 4 epochs. I removed this for now but clarify it if necessary.}

We compare BERT's behaviour with that of other pre-trained transformers such as RoBERTa \cite{liu_roberta_2019} and DeBERTa \cite{he2020deberta} fine-tuned with the same optimiser and hyper-parameters as above.
In order to also compare against non-transformer models, we report performance for a bi-LSTM-CRF \cite{lample2016neural} model with combined character-level and word-level representations. The model is comprised of 10 layers, with 300-dimensional word representations and 50-dimensional character representations, for a total of approximately 30 million trainable parameters. In our experiments, the model is trained with the Adam optimiser \cite{kingma2014adam} and a learning rate of $10^{-4}$ for 100 epochs using a CRF loss \cite{laffertyconditional}.

\section{Generalisation in noisy settings}
\label{sec:noise-effect}

\begin{figure}[bt]
    \centering
	\footnotesize
    \includegraphics[width=0.95\linewidth]{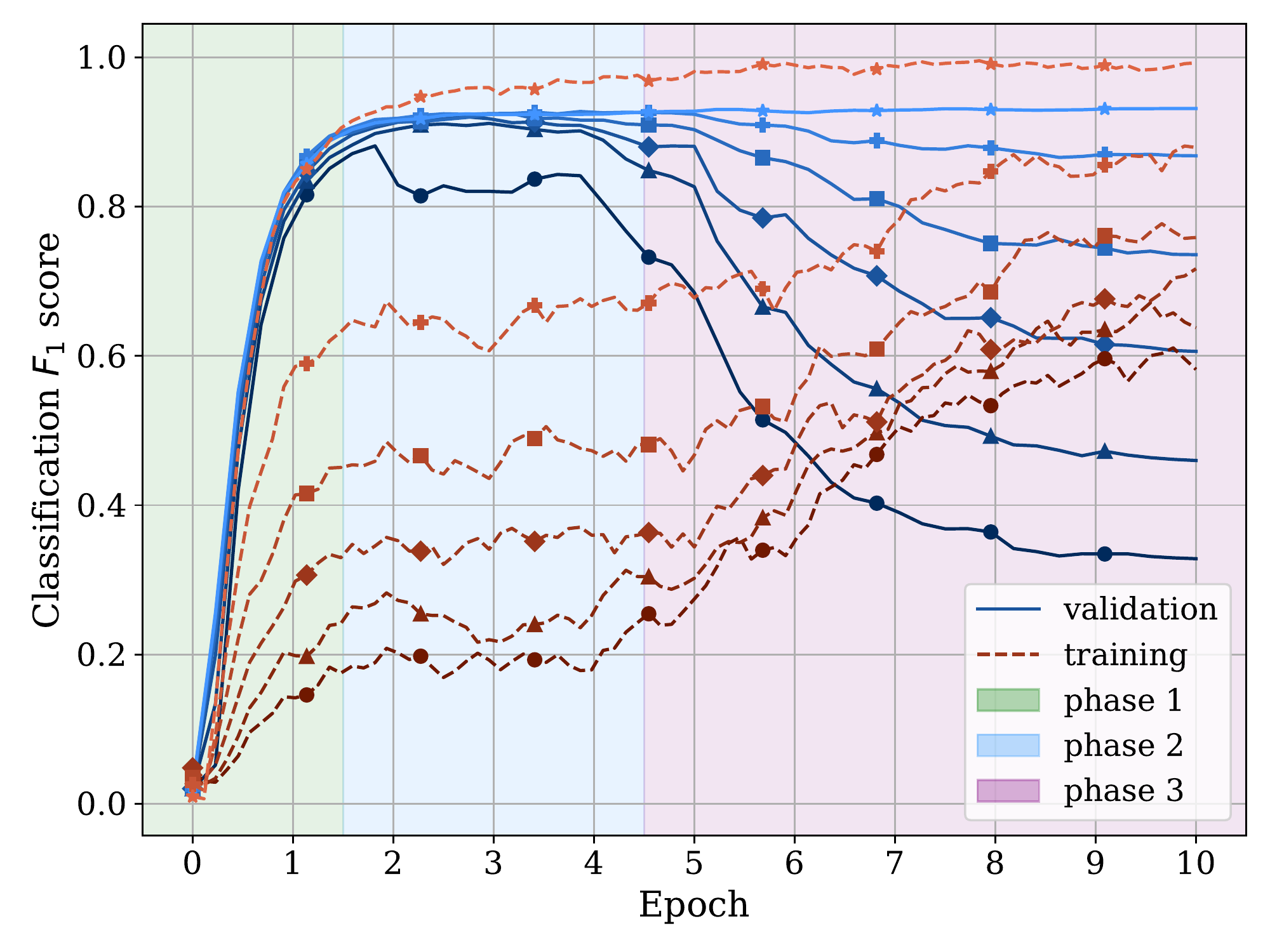}
    \caption{BERT performance (F$_1$) throughout the training process on the \texttt{CoNLL03} train and validation sets. Darker colours correspond to higher levels of noise (0\% to 50\%).}
    \label{fig:noise-performance-specific}
\end{figure}

We first investigate how BERT learns general patterns from datasets that contain label noise. Figure \ref{fig:noise-performance-specific} shows how the model performance on the \texttt{CoNLL03} training and validation sets changes when faced with varying levels of noise, from 0\% to 50\%.
Based on the progression of performance scores, we can divide BERT's learning process into roughly three distinct phases:

\begin{enumerate}[noitemsep]
	\item \textbf{Fitting}: The model uses the training data to learn how to generalise, effectively learning simple patterns that can explain as much of the training data as possible \cite{arpit_closer_2017}. Both the training and validation performance rapidly increase as the model learns these patterns.
	\item \textbf{Settling}: The increase in performance plateaus and neither the validation nor the training performance change considerably. The duration of this phase seems to be inversely proportional to the amount of noise present in the dataset.
	\item \textbf{Memorisation}: The model \bdemo{rapidly starts to memorise the noisy examples}, quickly improving the performance on training data while degrading the validation performance, effectively over-fitting to the noise in the dataset.
\end{enumerate}

\begin{figure}[bt]
    \centering
	\footnotesize
    \includegraphics[width=0.95\linewidth]{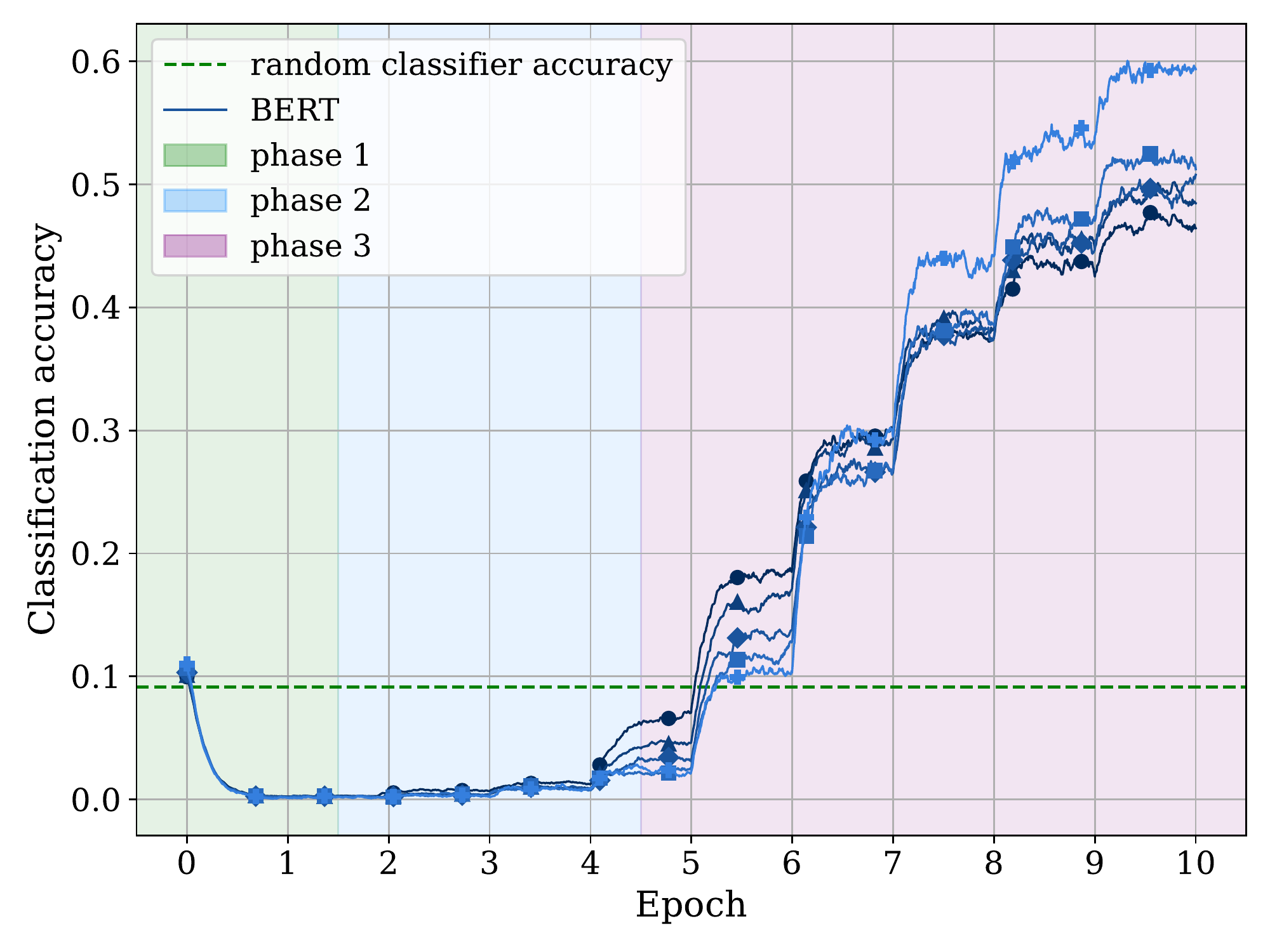}
    \caption{Classification accuracy of noisy examples in the training set for the \texttt{CoNLL03} dataset. Darker colours correspond to higher levels of noise (0\% to 50\%).}
    \label{fig:noise-accuracy}
\end{figure}

\paragraph{A second phase of learning} We find BERT to exhibit a distinct second \emph{settling} phase during which it does not over-fit. A resilience to label noise has been observed in other neural networks trained with gradient descent \cite{li2020gradient}. However, we find this phase to be much more prolonged in BERT compared to models pre-trained on other modalities such as a pre-trained ResNet fine-tuned on \texttt{CIFAR10}, which immediately starts memorising noisy examples (see Appendix \ref{app:second_phase_comparison} for a comparison). 
These results indicate that the precise point of early stopping is not as important when it comes to fine-tuning pre-trained language models. Similar optimal performance is retained for a substantial period, therefore training for a fixed number of epochs can be sufficient.

We illustrate BERT's behaviour by evaluating the token-level classification accuracy of noisy examples in Figure \ref{fig:noise-accuracy}. During the second phase, BERT completely ignores the noisy tokens and correctly misclassifies them, performing ``worse'' than a random classifier. The step-like improvements during the third stage show that the model is unable to learn any patterns from the noise and improves by repeatedly optimising on the same examples, gradually memorising them.

%We noted that the average gradient norm is at a minimum during the second phase but did not observe changes in prediction quality or confidence. We leave a more detailed analysis of BERT's behaviour in the second phase to future work.

\paragraph{Robustness to noise} We also observe in Figure \ref{fig:noise-performance-specific} that BERT is extremely robust to noise and over-fitting in general. In the absence of noise, the model does not over-fit and maintains its development set performance, regardless of the length of training. Even with a large proportion of noise, 
model performance comparable to training on the clean dataset can be achieved by stopping  the training process somewhere in the second phase.\footnote{Adding 30\% noise to the \texttt{CoNLL03} dataset causes only a 0.9\% decrease of validation performance in the second phase.}

We also hypothesise that due to the robustness to noise shown in the second phase of training, a noise detector can be constructed based only on BERT's training losses, without requiring any other information. We find that a simple detector that clusters the losses using k-means reliably achieves over 90\% noise-detection F$_1$ score in all our experiments, further showing how the model is able to actively detect and reject single noisy examples (see Appendix \ref{app:noise-detection} for details about the noise detection process).

\paragraph{Impact of pre-training} The above properties can mostly be attributed to BERT's pre-training process---after large-scale optimisation as a language model, the network is primed for learning general patterns and better able to ignore individual noisy examples. We find that a randomly initialised model with the same architecture does not only achieve lower overall performance but crucially does not exhibit's BERT's distinct second phase of learning and robustness to noise (see Appendix \ref{app:effect_of_pretraining}).

\paragraph{Other pre-trained transformers}
We also analyse the behaviour of other pre-trained transformers for comparison. Specifically, studying RoBERTa and DeBERTa, we find the same training pattern that was observed in BERT---all models show a clear division into the three phases described above. These models are also all very robust to label noise during the \emph{settling} phase of training. Notably, RoBERTa is even more resilient to label noise compared to the other two analysed models, despite DeBERTa outperforming it on public benchmarks \cite{he2020deberta}. Training and validation performance visualisations, such as those in Figure \ref{fig:noise-performance-specific}, can be found for both models in Appendix \ref{sec:other-transformers}.

\section{Forgetting of learned information}
\label{sec:forgetting}
Evaluating only the final model does not always provide the full picture regarding datapoint memorisation, as individual datapoints can be learned and forgotten multiple times during the training process.
%We now study to what extent BERT learns and forgets examples. 
Following \citet{toneva_empirical_2019}, we record a \textit{forgetting event} for an example at epoch $t$ if the model was able to classify it correctly at epoch $t-1$, but not at epoch $t$. Similarly, we identify a \textit{learning event} for an example at epoch $t$ if the model was not able to classify it correctly at epoch $t-1$, but it is able to do so at epoch $t$. A \textit{first learning event} thus happens at the first epoch when a model is able to classify an example correctly. We furthermore refer to examples with zero and more than zero forgetting events as \textit{unforgettable} and \textit{forgettable} examples, respectively, while the set of \textit{learned} examples includes all examples with one or more learning events.

In Table \ref{tab:bert-forgetting-events-short}, we show the number of forgettable, unforgettable, and learned examples on the training data of the \texttt{CoNLL03} and \texttt{JNLPBA} datasets for BERT, a non-pre-trained BERT, and a bi-LSTM model. We also show the ratio between forgettable and learned examples, which indicates how easily a model forgets learned information. We can observe that BERT forgets less than other models and that pre-training is crucial for retaining important information. We show the most forgettable examples in Appendix \ref{sec:forgettable_examples}, which tend to be atypical examples of the corresponding class.

\begin{table*}[bt]
	\centering
	\footnotesize
	\begin{tabular}{r|r|rrr|r} 
		\toprule
		\multicolumn{1}{r|}{\textbf{Dataset}} & 
		\multicolumn{1}{r|}{\textbf{Model}} & \multicolumn{1}{l}{\textbf{Forgettable} $N_f$} & \multicolumn{1}{l}{\textbf{Unforgettable} $N_u$} & \multicolumn{1}{l}{\textbf{Learned} $N_l$} & \multicolumn{1}{|l}{\textbf{$N_f/N_l$} (\%)}  \\ 
		\midrule
		\multirow{3}{*}{\texttt{CoNNL03}} & 
		bi-LSTM & 71.06\% & 29.94\% & 90.90\% & 78.17\% \\
		&non-pre-trained BERT & 9.89\% & 90.11\% & 99.87\% & 9.90\% \\
		&pre-trained BERT & 2.97\% & 97.03\% & 99.80\% & 2.98\% \\
		\midrule
		\multirow{3}{*}{\texttt{JNLPBA}} & 
		bi-LSTM & 97.16\% & 5.14\% & 98.33\% & 98.81\%\\
		&non-pre-trained BERT & 25.50\% & 74.50\% & 98.24\% & 25.96\% \\
		&pre-trained BERT & 16.62\% & 83.38\% & 98.18\% & 16.93\% \\
		\bottomrule
	\end{tabular}
	\caption{Number of forgettable, unforgettable, and learned examples during BERT training on the \texttt{CoNLL03} dataset and \texttt{JNLPBA} dataset.}
	\label{tab:bert-forgetting-events-short}
\end{table*}

\citet{toneva_empirical_2019} found that the number of forgetting events remains comparable across different architectures for the vision modality, given a particular dataset.\footnote{They report proportions of forgettable examples for MNIST, PermutedMNIST, CIFAR10, and CIFAR100 as 8.3\%, 24.7\%, 68.7\%, and 92.38\% respectively.} 
%%%% Compared to non-Transformer and non-pre-trained models, as well as models trained on images \cite{toneva_empirical_2019}, a pre-trained BERT model forgets examples it has learned at a dramatically lower rate.
However, our experiments show that the same does not necessarily hold for pre-trained language models. Specifically, there is a large discrepancy in the ratio between forgettable and learned examples for BERT ({\raise.17ex\hbox{$\scriptstyle\sim$}}3\%) and a bi-LSTM model ({\raise.17ex\hbox{$\scriptstyle\sim$}}80\%) on the \texttt{CoNLL03} dataset.

\begin{figure}[t]
    \centering
	\footnotesize
    \includegraphics[width=0.95\linewidth]{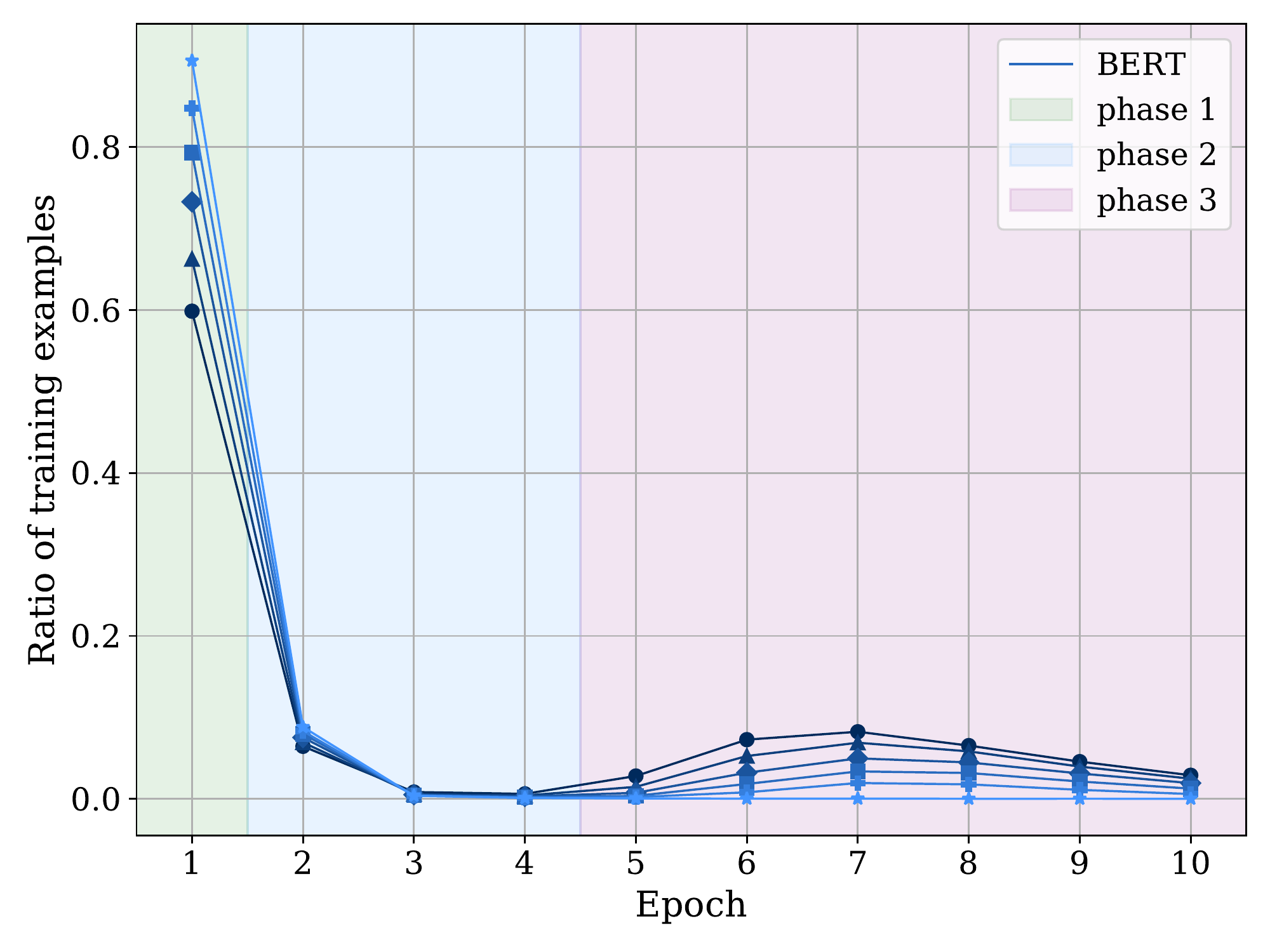}
    \caption{First learning events distribution during the training for various levels of noise on the \texttt{CoNLL03} dataset. Darker colours correspond to higher levels of noise (0\% to 50\%).}
    \label{fig:histogram-fle}
\end{figure}

We additionally analyse the distribution of first learning events throughout BERT's training on \texttt{CoNLL03} with label noise between 0\% and 50\% (Figure \ref{fig:histogram-fle}) and notice how BERT learns the majority of learned examples during the first epochs of training. As the training progresses, we see that BERT stops learning new examples entirely, regardless of the level of noise for the third and fourth epochs. Finally, in the last epochs BERT mostly memorises the noise in the data.\footnote{We conducted additional experiments on other datasets (see Appendix \ref{app:jnlpba_forgetting} for results on the \texttt{JNLPBA} dataset). In all cases we observe the same distribution of first learning events throughout training.}

\section{BERT in low-resource scenarios}\label{sec:few-shot}
In the previous sections, we have observed that BERT learns examples and generalises very early in training. We will now examine if the same behaviour applies in low-resource scenarios where a minority class is only observed very few times. To this end, we remove from the \texttt{CoNLL03} training set all sentences containing tokens with the minority labels \texttt{MISC} and \texttt{LOC} except for a predetermined number of such sentences. We repeat the process for the \texttt{JNLPBA} dataset with the \texttt{DNA} and \texttt{Protein} labels.

\begin{figure}[bt]
	\centering
	\footnotesize
	\includegraphics[width=0.95\linewidth]{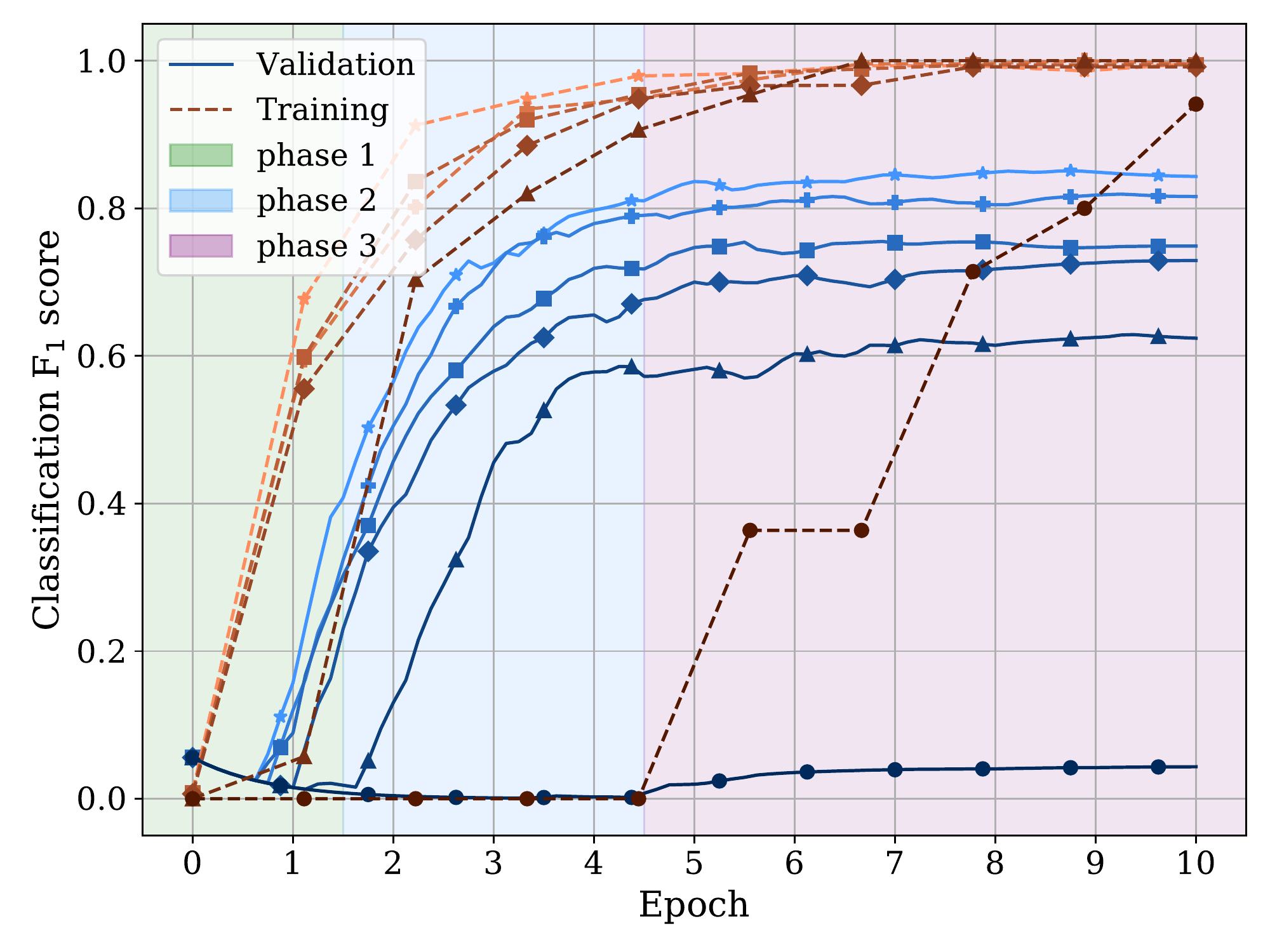}
    \caption{BERT performance (F$_1$) throughout the training process on the \texttt{CoNLL03} dataset with varying number of sentences containing the \texttt{LOC} class. Darker colours correspond to fewer examples of the \texttt{LOC} class available (5 to 95 in steps of 20).}
	\label{fig:noisy-f1-score-loc-03}
\end{figure}

\begin{figure}[bt]
	\centering
	\footnotesize
	\includegraphics[width=0.95\linewidth]{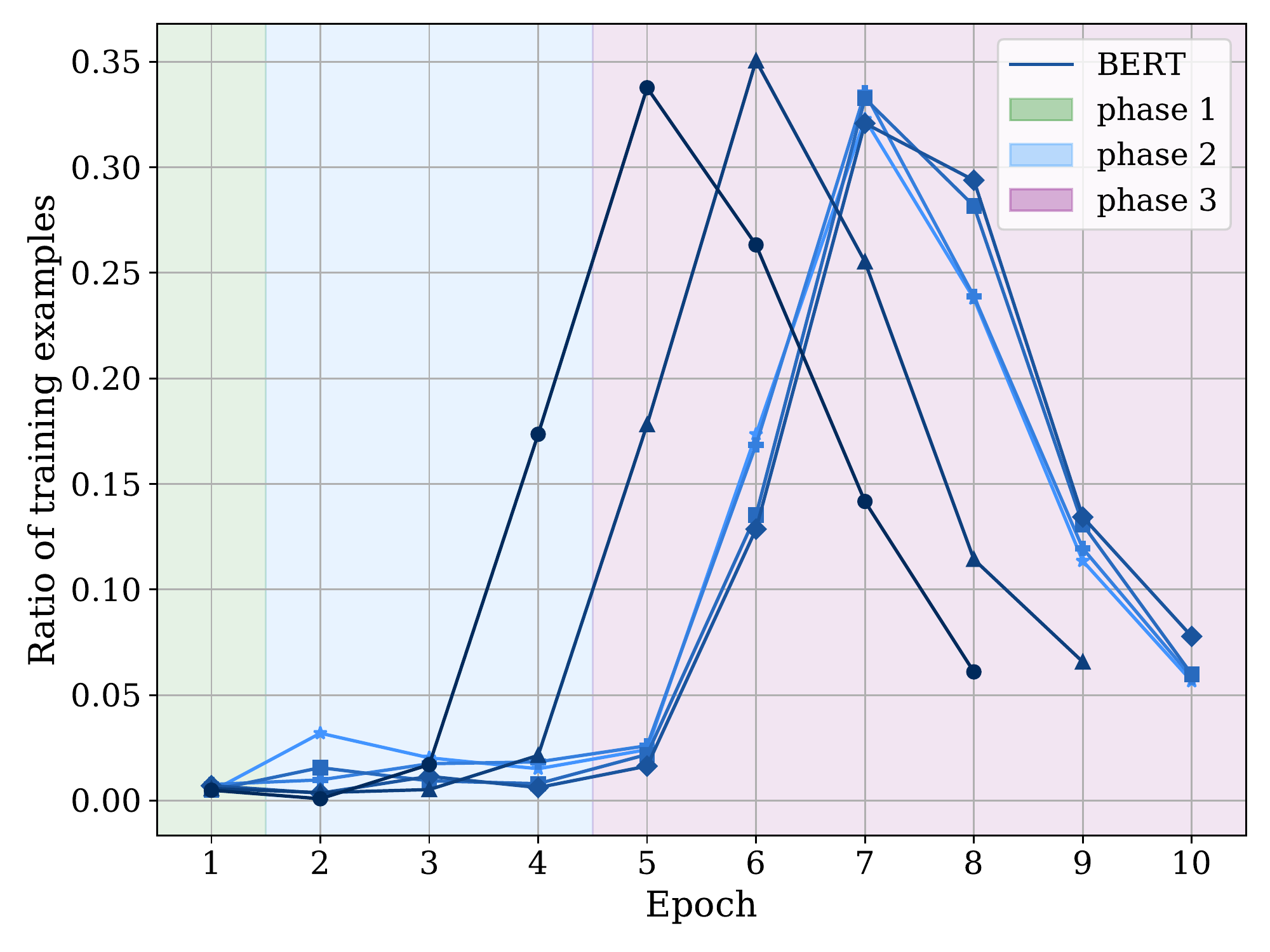}
    \caption{First learning events distribution during the training on the \texttt{CoNLL03} dataset with varying number of sentences containing the \texttt{LOC} class. Darker colours correspond to fewer examples of the \texttt{LOC} class available (5 to 95 in steps of 20).}
	\label{fig:fles-loc-03}
\end{figure}

We conduct similar experiments to the previous sections by
% introducing a fixed amount of label noise and
studying how different numbers of sentences containing the target class affect BERT's ability to learn and generalise. We report in Figure \ref{fig:noisy-f1-score-loc-03} the training and validation classification F$_1$ score for the \texttt{CoNLL03} datasets from which all but few (5 to 95) sentences containing the \texttt{LOC} label were removed. Note that the reported performance in this experiment refers to the \texttt{LOC} class only. In Figure \ref{fig:fles-loc-03} we also report the distribution of first learning events for the \texttt{LOC} class in the same setting. Two phenomena can be observed: 1) reducing the number of sentences greatly reduces the model's ability to generalise (validation performance decreases yet training performance remains comparable); and 2) when fewer sentences are available, they tend to be learned in earlier epochs for the first time. Corresponding experiments on the \texttt{MISC} label can be found in Appendix \ref{app:misc-experiments}.
	
\begin{figure}[bt]
	\centering
	\footnotesize
	\includegraphics[width=0.95\linewidth]{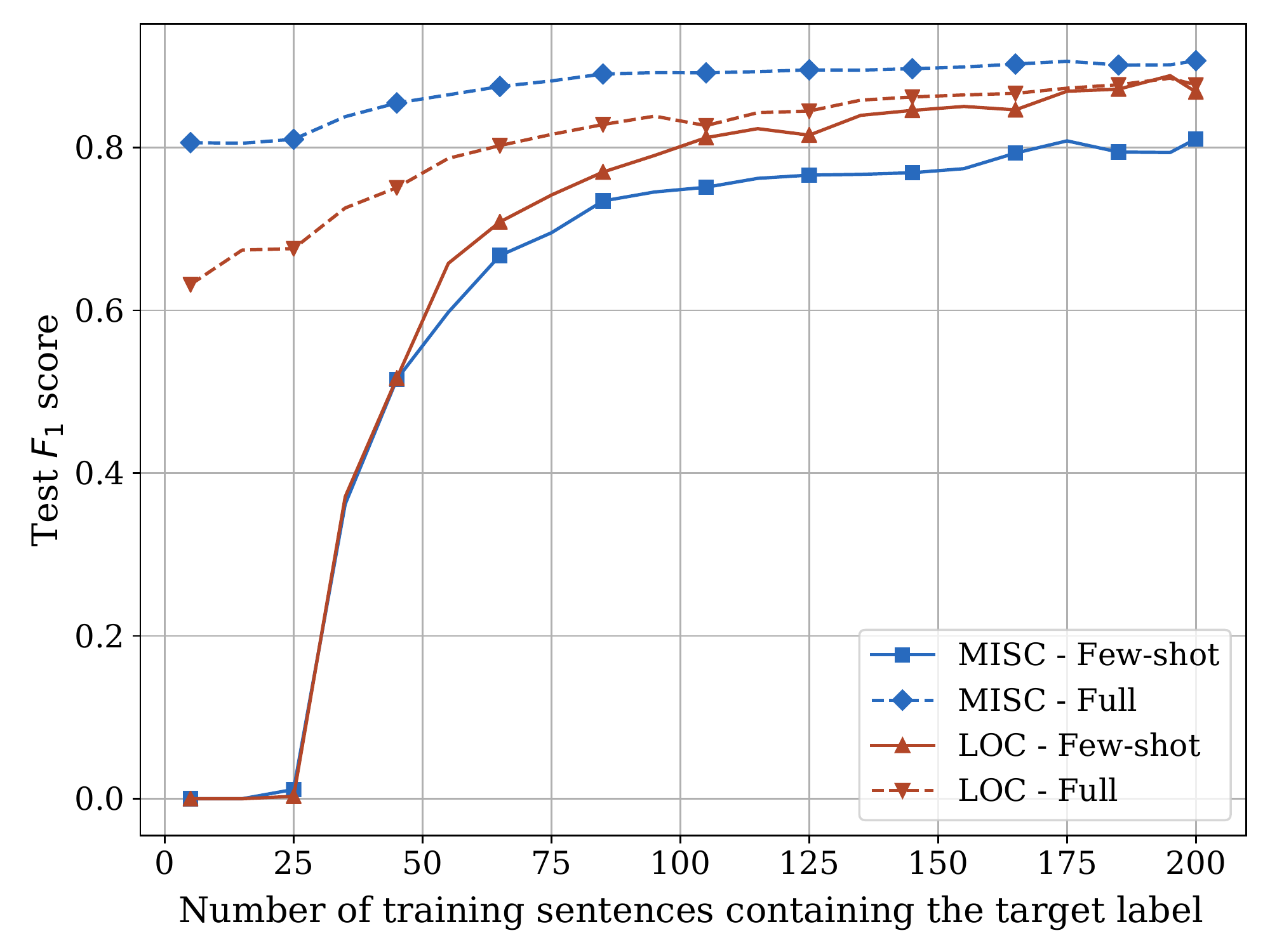}
	\caption{BERT final validation entity-level F$_1$ score on the few-shot class keeping varying numbers of sentences containing examples of a selected class on the \texttt{CoNLL03} dataset.}
	\label{fig:reduced-sentences-final}
\end{figure}

\begin{figure}[bt]
	\centering
	\footnotesize
	\includegraphics[width=0.95\linewidth]{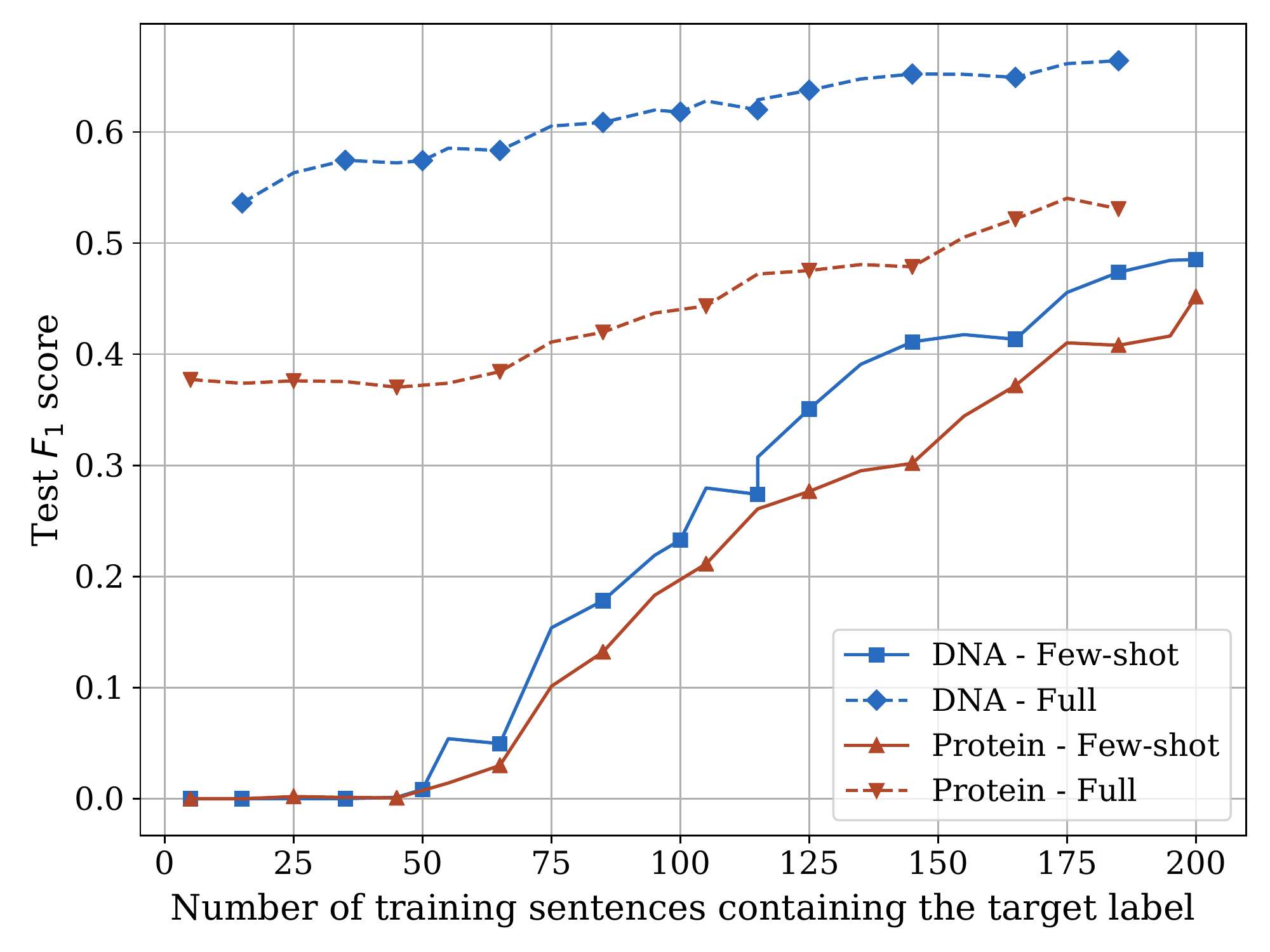}
	\caption{BERT final validation entity-level F$_1$ score on the few-shot class keeping varying numbers of sentences containing examples of a selected class on the \texttt{JNLPBA} dataset.}
	\label{fig:jnlpba-reduced-sentences-final-paper}
\end{figure}

We also show the average entity-level F$_1$ score on tokens belonging to the minority label and the model performance for the full NER task (i.e. considering all classes) for the \texttt{CoNLL03} and \texttt{JNLPBA} datasets in Figures \ref{fig:reduced-sentences-final} and \ref{fig:jnlpba-reduced-sentences-final-paper} respectively. For the \texttt{CoNLL03} dataset, we observe that BERT needs at least 25 examples of a minority label in order to be able to start learning it. Performance rapidly improves from there and plateaus at around 100 examples. For the \texttt{JNLPBA} dataset, the minimum number of examples increases to almost 50 and the plateau occurs for a higher number of examples.
On the challenging \texttt{WNUT17} dataset, BERT achieves only 44\% entity-level F$_1$. This low performance is attributable to the absence of entity overlap between training set and test set, which increases the inter-class variability of the examples.

\section{ProtoBERT for few-shot learning}\label{sec:solution}

\begin{figure*}[htb]
	\centering
	\footnotesize
	\includegraphics[width=0.9\linewidth]{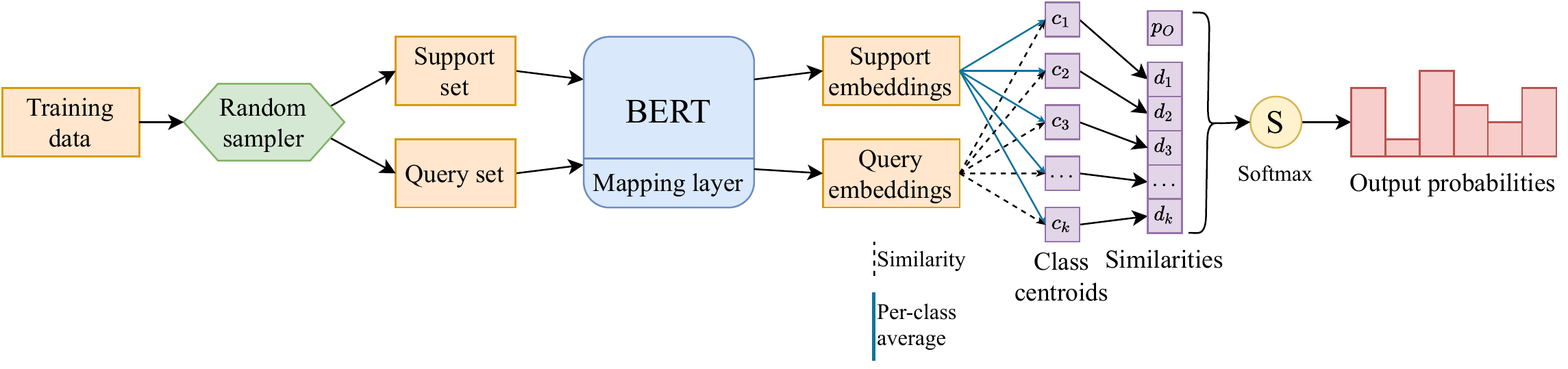}
	\caption{Schematic representation of the inference using a BERT model with a prototypical network layer.}
	\label{fig:ProtoBERT-arch}
\end{figure*}

In order to address BERT's limitations in few-shot learning, we propose a new model, ProtoBERT that combines BERT's pre-trained knowledge with the few-shot capabilities of prototypical networks \cite{snell_prototypical_2017} for sequence labelling problems. The method builds an embedding space where the inputs are clustered on a per-class basis, allowing us to classify a token by finding its closest centroid and assigning it the corresponding class.
The model can be seen in Figure \ref{fig:ProtoBERT-arch}.

We first define a support set $S$, which we use as context for the classification and designate with $S_k$ all elements of $S$ that have label $k$. We refer to the set of points that we want to classify as the query set $Q$, with $l(Q_i)$ indicating the label of the $i^{\text{th}}$ element in $Q$.  We will also refer to $f$ as the function computed by BERT augmented with a linear layer, which produces an $M$ dimensional output.
	
The model then classifies a given input $\mathbf{x}$ as follows: for each class $k$, we compute the centroid of the class in the learned feature space as the mean of all the elements that belong to class $k$ in the support set $S$:
	\begin{equation}\label{eq:centroid}
		\mathbf{c}_k = \frac{1}{|S_k|} \sum_{\mathbf{x}_i \in S_k}f(\mathbf{x}_i)
	\end{equation}
Then, we compute the distance from each input $\mathbf{x} \in Q$ to each centroid:
	\begin{equation*}
		dist_k = d(f(\mathbf{x}), \mathbf{c}_k)
	\end{equation*}
	and collect them in a vector $v \in \mathbb{R}^k$.
Finally, we compute the probability of $\mathbf{x}$ belonging to class $k$ as
	\begin{align*}
		p(y=k \mid \mathbf{x} ) &=\frac{\exp \left(-d\left(f( \mathbf{x} ), \mathbf{c} _{k}\right)\right)}{\sum_{k^{\prime}} \exp \left(-d\left(f( \mathbf{x} ), \mathbf{c} _{k^{\prime}}\right)\right)} =\\&= softmax(-v)_k
	\end{align*}

The model is trained by optimising the cross-entropy loss between the above probability and the one-hot ground-truth label of $\mathbf{x}$. Crucially, $S$ and $Q$ are not a fixed partition of the training set but change at each training step. Following \citet{snell_prototypical_2017}, we use Euclidean distance as a choice for the function $d$. 

In order to take into account the extreme under-representation of some classes, we create the support by sampling $s_1$ elements from each minority class and $s_2$ elements from each non-minority class. A high ratio $s_1/s_2$ gives priority to the minority classes, while a low ratio puts more emphasis on the other classes. We then similarly construct the query set with a fixed ratio $n$ between the minority classes and the non-minority classes.

% In order to take into account the extreme under-representation of some classes, we propose to sample the support set $S$ and query set $Q$ at every training step as follows: We first randomly select $s_1$ examples of the training set $X$ belonging to each minority class, add them to $S$, and add the remaining examples of the minority classes to $Q$. We then select $s_2$ examples from $X$ not belonging to the minority class and also add them to $S$. Finally, we select $n \cdot s_1 \cdot c_m$ examples from $X$ not already either in $S$ or $Q$ and add them to $Q$ where $n$ is the ratio of negative samples and $c_m$ is the number of minority classes in the training set.

% This results in a support set $S$ containing $c_m \cdot s_1 + s_2$ examples and a query set $Q$ containing $\sum_{k \in C_m}\left(|X_k| - s_1\right) + n \cdot s_1 \cdot c_m$ examples where $C_m$ is the set of minority labels and $X_k$ is the number of examples in the training set belonging to label $k$. Overall, a high ratio $s_1/s_2$ gives priority to the minority classes, while a low ratio puts more emphasis on the other classes.

% \todo{MR: Too vague, can't just leave it like this. Need to say a bit more, e.g. "the minority class is sampled x times more frequently for the support set" or something. Or put the details in the appendix.}

For NER, rather than learning a common representation for the negative class ``\texttt{O}'', we only want the model to treat it as a fallback when no other similar class can be found. For this reason, we define the vector of distances $v$ as follows:
\begin{equation*}
	v = (d_O,\ dist_0,\ \ldots,\ dist_k)
\end{equation*}
where $d_O$ is a scalar parameter of the network that is trained along with the other parameters. Intuitively, we want to classify a point as a \textit{non-entity} (i.e. class \texttt{O}) when it is not close enough to any centroid, where $d_O$ represents the threshold for which we consider a point ``close enough''.

% In order to take into account the extreme under-representation of some classes, we propose to sample the support set $S$ and query set $Q$ at every training step as follows: We first randomly select $s_1$ examples of the training set $X$ belonging to each minority class, add them to $S$, and add the remaining examples of the minority classes to $Q$. We then select $s_2$ examples from $X$ not belonging to the minority class and also add them to $S$. Finally, we select $n \cdot s_1 \cdot c_m$ examples from $X$ not already either in $S$ or $Q$ and add them to $Q$ where $n$ is the ratio of negative samples and $c_m$ is the number of minority classes in the training set.

% This results in a support set $S$ containing $c_m \cdot s_1 + s_2$ examples and a query set $Q$ containing $\sum_{k \in C_m}\left(|X_k| - s_1\right) + n \cdot s_1 \cdot c_m$ examples where $C_m$ is the set of minority labels and $X_k$ is the number of examples in the training set belonging to label $k$. Overall, a high ratio $s_1/s_2$ gives priority to the minority classes, while a low ratio puts more emphasis on the other classes.

% \todo{MR: These last two paragraphs seem unnecessarily complex. If you need space (currently there's a figure on the 9th page), you could say something like "In order to take into account the extreme under-representation of some classes, we perform weighted sampling, putting more focus on the minority class". And move all the rest of this sampling detail into the appendix.}

If no example of a certain class is available in the support set during the training, we assign a distance of $400$, making it effectively impossible to mistakenly classify the input as the missing class during that particular batch. Finally, we propose two ways to compute the class of a token at test time. The first method employs all examples from $X$ to calculate the centroids needed at test time, which produces better results but is computationally expensive for larger datasets.

The second method approximates the centroid $\mathbf{c}_k$ using the moving average of the centroids produced at each training step:
\begin{equation*}
		\mathbf{c}_k^{(t)} \leftarrow \alpha\ \mathbf{c}_k^{(t)} \cdot (1 - \alpha)\ \mathbf{c}_k^{(t-1)}
\end{equation*}
where $\alpha$ is a weighting factor. This method results in little overhead during training and only performs marginally worse than the first method.

\subsection{Experimental results}
\begin{figure}[bt]
	\centering
	\footnotesize
	\includegraphics[width=0.95\linewidth]{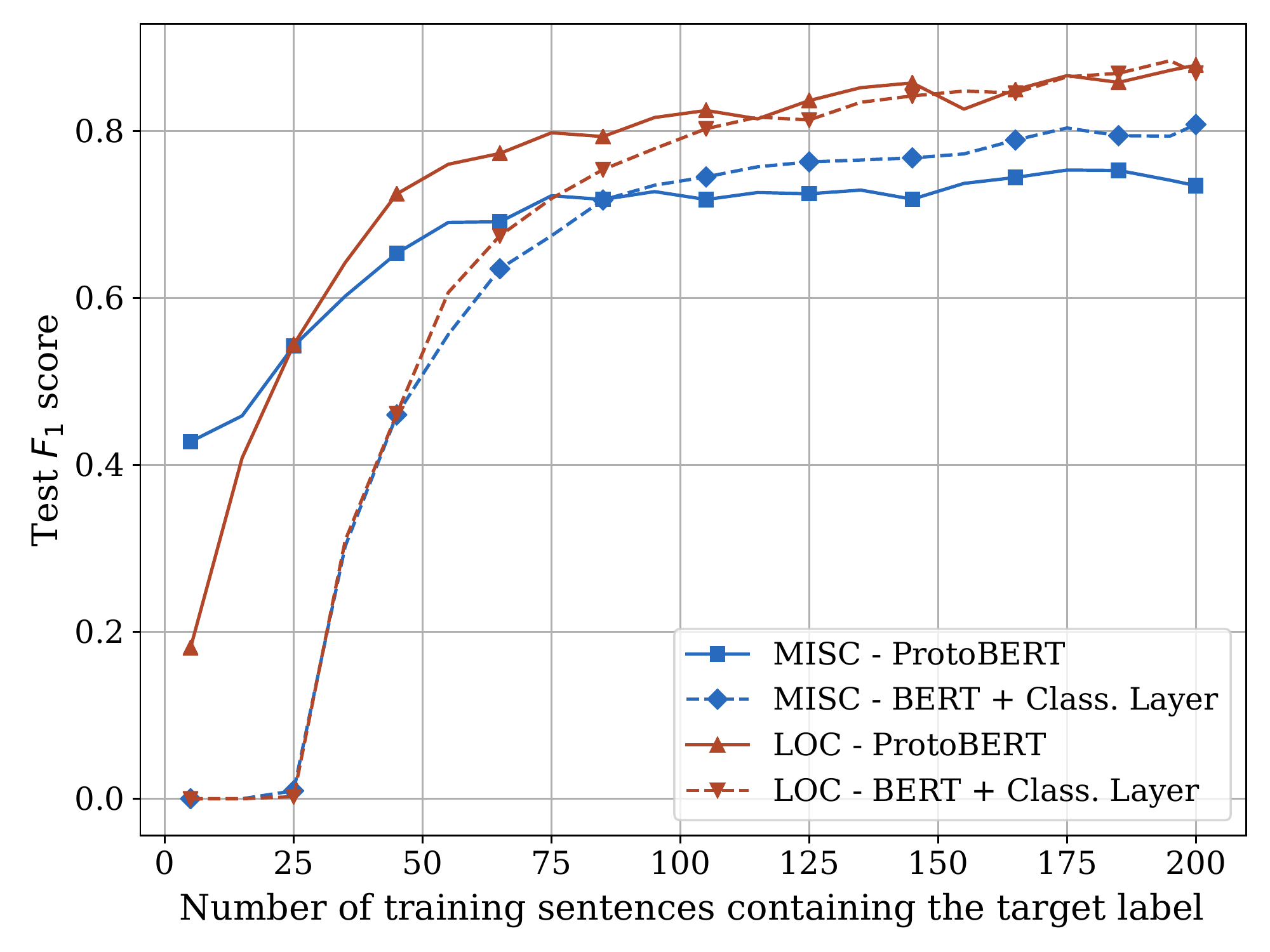}
	\caption{Model performance comparison between the baseline model and ProtoBERT for the \texttt{CoNLL03} dataset, reducing the sentences containing the \texttt{MISC} and \texttt{LOC} classes. Results reported as F$_1$ score on the few-shot classes.}
	\label{fig:conll03-proto-min}
\end{figure}

\begin{figure}[bt]
	\centering
    \footnotesize  
	\includegraphics[width=0.95\linewidth]{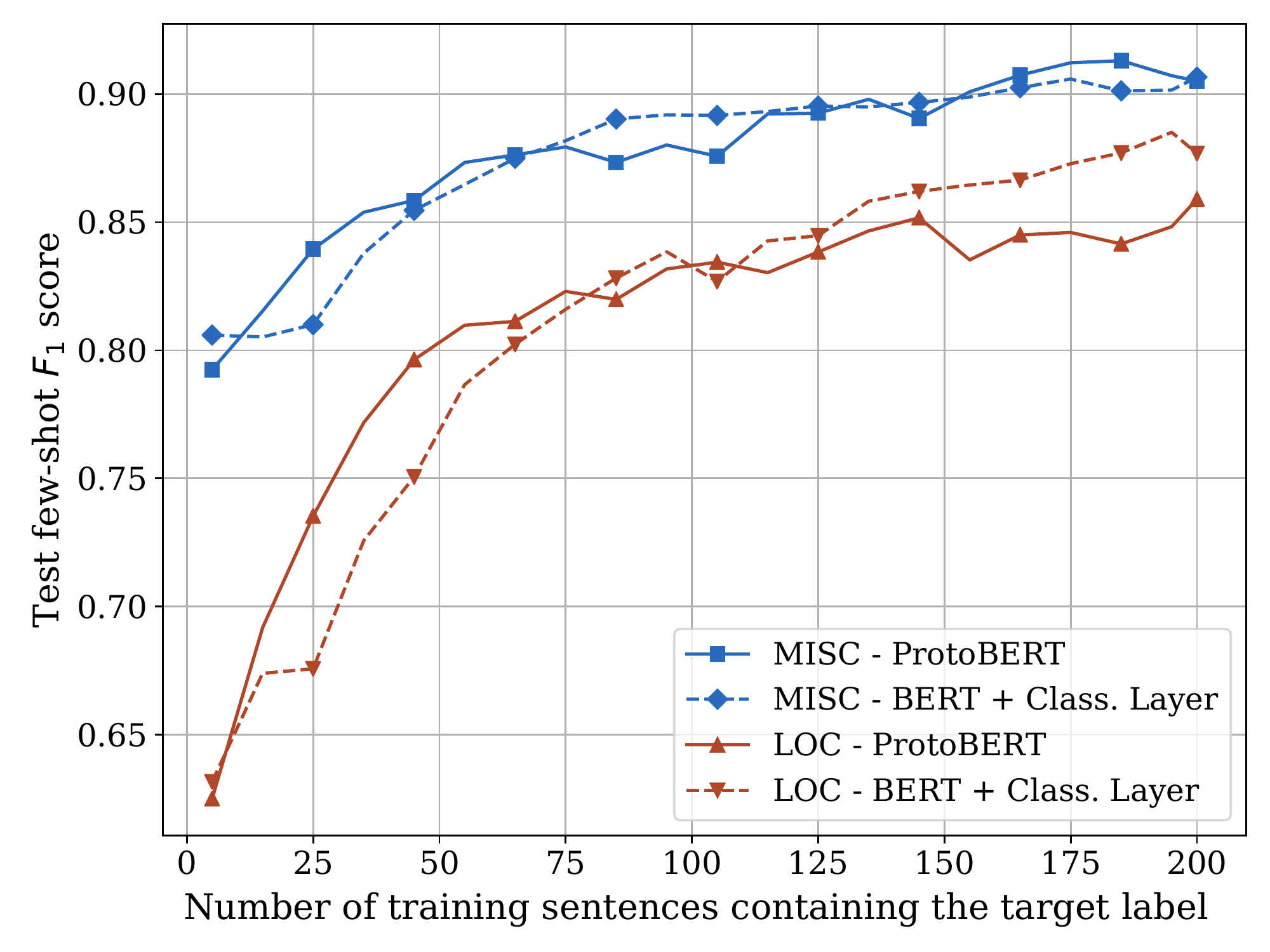}
	\caption{Model performance comparison between the baseline model and ProtoBERT for the \texttt{CoNLL03} dataset, reducing the sentences containing the \texttt{MISC} and \texttt{LOC} class. Results reported as F$_1$ score on all classes.}
	\label{fig:conll03-proto-full}
\end{figure}

We first compare ProtoBERT to the standard pre-trained BERT model with a classification layer on the \texttt{CoNLL03} and \texttt{JNLPBA} datasets with a smaller number of sentences belonging to the minority classes. We show the results on the few-shot classes and for the full dataset for \texttt{CoNLL03} in Figures \ref{fig:conll03-proto-min} and \ref{fig:conll03-proto-full} respectively. Similarly, we show the results for the few-shot class for \texttt{JNLPBA} in Figure \ref{fig:jnlpba-proto-min-paper-final}.\footnote{A comparison on the full classification task can be found in Appendix \ref{app:proto-jnlpba}.} In all cases ProtoBERT consistently surpasses the performance of the baseline when training on few examples of the minority class. It particularly excels in the extreme few-shot setting, e.g. outperforming BERT by 40 F$_1$ points with 15 sentences containing the \texttt{LOC} class.
As the number of available examples of the minority class increases, BERT starts to match ProtoBERT's performance and outperforms it on the full dataset in some cases.

\begin{figure}[bt]
	\centering
	\footnotesize
	\includegraphics[width=0.95\linewidth]{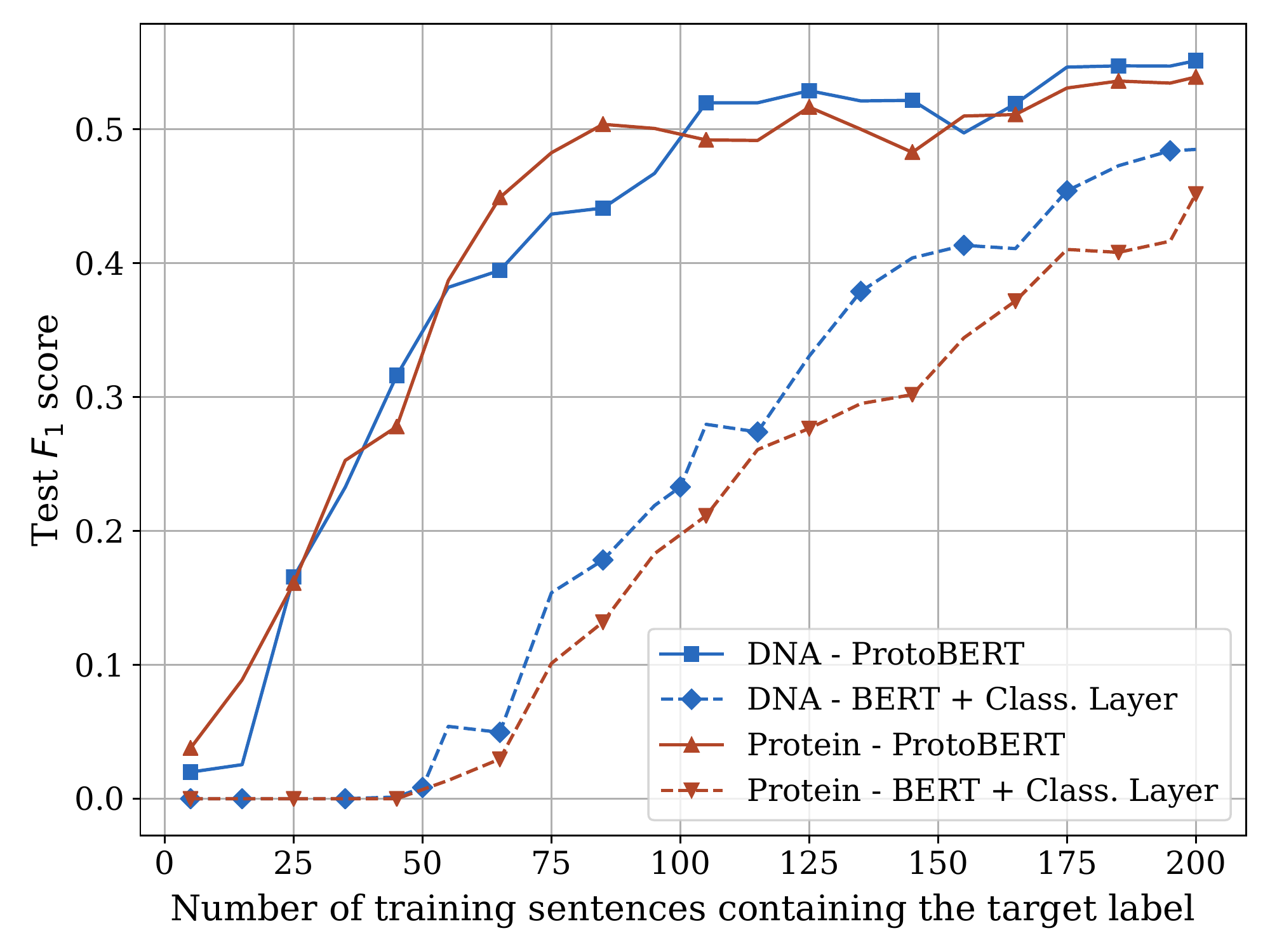}
	\caption{Model performance comparison between the baseline model and ProtoBERT for the \texttt{JNLPBA} dataset, reducing the sentences containing the \texttt{DNA} and \texttt{Protein} classes. Results reported as F$_1$ score on the few-shot classes.}
	\label{fig:jnlpba-proto-min-paper-final}
\end{figure}

\begin{table*}[htb]
    \centering
	\footnotesize
    \begin{tabular}{l|rrr}
        \toprule
        \textbf{Model}                         & \textbf{CoNLL03}                & \textbf{JNLPBA}                 & \textbf{WNUT17}                 \\ \midrule
        State of the art                       & 93.50                           & 77.59                           & 50.03                           \\
        BERT + classification layer (baseline) & 89.35                           & \textbf{75.36} & 44.09                           \\ \midrule
        ProtoBERT                              & \textbf{89.87} & 73.91                           & \textbf{48.62} \\
        ProtoBERT + running centroids          & 89.46                           & 73.54                           & 48.56                           \\ \bottomrule
    \end{tabular}
    \caption{Comparison between the baseline model, the current state-of-the-art\footnotemark and the proposed architecture on the \texttt{CoNLL03}, \texttt{JNLPBA} and \texttt{WNUT17} datasets evaluated using entity-level F$_1$ score. The state of the art is \citet{baevski2019cloze}, \citet{lee_biobert_2019}, and \citet{wang_crossweigh_2019} respectively.}
    \label{tab:base-vs-proto}
\end{table*}

While the main strength of ProtoBERT is on few-shot learning, we evaluate it also on the full \texttt{CoNLL03}, \texttt{JNLPBA} and \texttt{WNUT17} datasets (without removing any sentences) in Table \ref{tab:base-vs-proto}. In this setting, the proposed architecture achieves results mostly similar to the baseline while considerably outperforming it on the \texttt{WNUT17} dataset of rare entities.

The results in this section show that ProtoBERT, while designed for few-shot learning, performs at least on par with its base model in all tasks. This allows the proposed model to be applied to a much wider range of tasks and datasets without negatively affecting the performance if no label imbalance is present, while bringing a substantial improvement in few-shot scenarios.

We conduct an ablation study to verify the effect of our improved centroid computation method. From the results in Table \ref{tab:base-vs-proto} we can affirm that, while a difference in performance does exist, it is quite modest (0.1--0.4\%). On the other hand, this method reduces the training time and therefore energy consumption \cite{strubell-etal-2019-energy} to one third of the original method on \texttt{CoNLL03} and we expect the reduction to be even greater for larger datasets.

\section{Conclusion}
In this study, we investigated the learning process during fine-tuning of pre-trained language models, focusing on generalisation and memorisation.
By formulating  experiments that  allow  for  full  control over the label distribution in the training data, we study the learning dynamics of the models in conditions of high label noise and low label frequency.
The experiments show that BERT is capable of reaching near-optimal performance even when a large proportion of the training set labels has been corrupted. We find that this ability is due to the model's tendency to separate the training into three distinct phases: fitting, settling, and memorisation, which allows the model to ignore noisy examples in the earlier epochs. The pretrained models experience a prolonged settling phase when fine-tuned, during which their performance remains optimal, indicating that the precise area of early stopping is less crucial.

Furthermore, we show that the number of available examples greatly affects the learning process, influencing both when the examples are memorised and the quality of the generalisation. We show that BERT fails to learn from examples in extreme few-shot settings, completely ignoring the minority class at test time. To overcome this limitation, we augment BERT with a prototypical network. This approach partially solves the model's limitations by enabling it to perform well in extremely low-resource scenarios and also achieves comparable performance in higher-resource settings.

\section*{Acknowledgements}

Michael is funded by the UKRI CDT in AI for Healthcare\footnote{\url{http://ai4health.io}} (Grant No. P/S023283/1).

% \clearpage
\bibliography{anthology,bib, custom}

\begin{thebibliography}{36}
\expandafter\ifx\csname natexlab\endcsname\relax\def\natexlab#1{#1}\fi

\bibitem[{Arpit et~al.(2017)Arpit, Jastrzębski, Ballas, Krueger, Bengio,
  Kanwal, Maharaj, Fischer, Courville, Bengio, and
  Lacoste-Julien}]{arpit_closer_2017}
Devansh Arpit, Stanisław Jastrzębski, Nicolas Ballas, David Krueger, Emmanuel
  Bengio, Maxinder~S. Kanwal, Tegan Maharaj, Asja Fischer, Aaron Courville,
  Yoshua Bengio, and Simon Lacoste-Julien. 2017.
\newblock \href {http://arxiv.org/abs/1706.05394} {A {Closer} {Look} at
  {Memorization} in {Deep} {Networks}}.
\newblock \emph{arXiv:1706.05394 [cs, stat]}.
\newblock ArXiv: 1706.05394.

\bibitem[{Augenstein et~al.(2017)Augenstein, Derczynski, and
  Bontcheva}]{augenstein_generalisation_2017}
Isabelle Augenstein, Leon Derczynski, and Kalina Bontcheva. 2017.
\newblock \href {http://arxiv.org/abs/1701.02877} {Generalisation in {Named}
  {Entity} {Recognition}: {A} {Quantitative} {Analysis}}.
\newblock \emph{arXiv:1701.02877 [cs]}.
\newblock ArXiv: 1701.02877.

\bibitem[{Baevski et~al.(2019)Baevski, Edunov, Liu, Zettlemoyer, and
  Auli}]{baevski2019cloze}
Alexei Baevski, Sergey Edunov, Yinhan Liu, Luke Zettlemoyer, and Michael Auli.
  2019.
\newblock \href {https://doi.org/10.18653/v1/D19-1539} {Cloze-driven
  pretraining of self-attention networks}.
\newblock In \emph{Proceedings of the 2019 Conference on Empirical Methods in
  Natural Language Processing and the 9th International Joint Conference on
  Natural Language Processing (EMNLP-IJCNLP)}, pages 5360--5369, Hong Kong,
  China. Association for Computational Linguistics.

\bibitem[{Carlini et~al.(2019)Carlini, Liu, Erlingsson, Kos, and
  Song}]{carlini_secret_2019}
Nicholas Carlini, Chang Liu, Úlfar Erlingsson, Jernej Kos, and Dawn Song.
  2019.
\newblock \href {http://arxiv.org/abs/1802.08232} {The {Secret} {Sharer}:
  {Evaluating} and {Testing} {Unintended} {Memorization} in {Neural}
  {Networks}}.
\newblock \emph{arXiv:1802.08232 [cs]}.
\newblock ArXiv: 1802.08232.

\bibitem[{Collier and Kim(2004)}]{collier_introduction_2004}
Nigel Collier and Jin-Dong Kim. 2004.
\newblock \href {https://www.aclweb.org/anthology/W04-1213} {Introduction to
  the {Bio}-entity {Recognition} {Task} at {JNLPBA}}.
\newblock In \emph{Proceedings of the {International} {Joint} {Workshop} on
  {Natural} {Language} {Processing} in {Biomedicine} and its {Applications}
  ({NLPBA}/{BioNLP})}, pages 73--78, Geneva, Switzerland. COLING.

\bibitem[{Deng et~al.(2009)Deng, Dong, Socher, Li, Li, and
  Fei-Fei}]{deng2009imagenet}
Jia Deng, Wei Dong, Richard Socher, Li-Jia Li, Kai Li, and Li~Fei-Fei. 2009.
\newblock \href {https://ieeexplore.ieee.org/document/5206848} {Imagenet: A
  large-scale hierarchical image database}.
\newblock In \emph{2009 {IEEE} conference on computer vision and pattern
  recognition}, pages 248--255. {IEEE}.

\bibitem[{Derczynski et~al.(2017)Derczynski, Nichols, van Erp, and
  Limsopatham}]{derczynski_results_2017}
Leon Derczynski, Eric Nichols, Marieke van Erp, and Nut Limsopatham. 2017.
\newblock \href {https://doi.org/10.18653/v1/W17-4418} {Results of the
  {WNUT2017} {Shared} {Task} on {Novel} and {Emerging} {Entity} {Recognition}}.
\newblock In \emph{Proceedings of the 3rd {Workshop} on {Noisy}
  {User}-generated {Text}}, pages 140--147, Copenhagen, Denmark. Association
  for Computational Linguistics.

\bibitem[{Devlin et~al.(2019)Devlin, Chang, Lee, and
  Toutanova}]{devlin_bert_2019}
Jacob Devlin, Ming-Wei Chang, Kenton Lee, and Kristina Toutanova. 2019.
\newblock \href {http://arxiv.org/abs/1810.04805} {{BERT}: {Pre}-training of
  {Deep} {Bidirectional} {Transformers} for {Language} {Understanding}}.
\newblock \emph{arXiv:1810.04805 [cs]}.
\newblock ArXiv: 1810.04805.

\bibitem[{He et~al.(2015)He, Zhang, Ren, and Sun}]{he_deep_2015}
Kaiming He, Xiangyu Zhang, Shaoqing Ren, and Jian Sun. 2015.
\newblock \href {http://arxiv.org/abs/1512.03385} {Deep {Residual} {Learning}
  for {Image} {Recognition}}.
\newblock \emph{arXiv:1512.03385 [cs]}.
\newblock ArXiv: 1512.03385.

\bibitem[{He et~al.(2020)He, Liu, Gao, and Chen}]{he2020deberta}
Pengcheng He, Xiaodong Liu, Jianfeng Gao, and Weizhu Chen. 2020.
\newblock \href {https://arxiv.org/abs/2006.03654} {{DeBERTa: Decoding-enhanced
  BERT with Disentangled Attention}}.
\newblock \emph{arXiv e-prints}, pages arXiv--2006.

\bibitem[{Hendrycks et~al.(2020)Hendrycks, Liu, Wallace, Dziedzic, Krishnan,
  and Song}]{hendrycks_pretrained_2020}
Dan Hendrycks, Xiaoyuan Liu, Eric Wallace, Adam Dziedzic, Rishabh Krishnan, and
  Dawn Song. 2020.
\newblock \href {http://arxiv.org/abs/2004.06100} {Pretrained {Transformers}
  {Improve} {Out}-of-{Distribution} {Robustness}}.
\newblock \emph{arXiv:2004.06100 [cs]}.
\newblock ArXiv: 2004.06100.

\bibitem[{Howard and Ruder(2018)}]{Howard2018ulmfit}
Jeremy Howard and Sebastian Ruder. 2018.
\newblock \href {http://arxiv.org/abs/1801.06146} {{Universal Language Model
  Fine-tuning for Text Classification}}.
\newblock In \emph{Proceedings of ACL 2018}.

\bibitem[{Jawahar et~al.(2019)Jawahar, Sagot, and Seddah}]{jawahar2019does}
Ganesh Jawahar, Beno{\^\i}t Sagot, and Djam{\'e} Seddah. 2019.
\newblock \href {https://aclanthology.org/P19-1356.pdf} {{What Does BERT Learn
  about the Structure of Language?}}
\newblock In \emph{Proceedings of the 57th Annual Meeting of the Association
  for Computational Linguistics}, pages 3651--3657.

\bibitem[{Kingma and Ba(2014)}]{kingma2014adam}
Diederik~P Kingma and Jimmy Ba. 2014.
\newblock \href
  {https://ui.adsabs.harvard.edu/abs/2014arXiv1412.6980K/abstract} {Adam: A
  method for stochastic optimization}.
\newblock \emph{arXiv e-prints}, pages arXiv--1412.

\bibitem[{Krizhevsky(2009)}]{krizhevsky_learning_2009}
Alex Krizhevsky. 2009.
\newblock \href
  {https://citeseerx.ist.psu.edu/viewdoc/download?doi=10.1.1.222.9220&rep=rep1&type=pdf}
  {Learning {Multiple} {Layers} of {Features} from {Tiny} {Images}}.
\newblock \emph{University of Toronto}.

\bibitem[{Kumar et~al.(2020)Kumar, Makhija, and Gupta}]{kumar2020user}
Ankit Kumar, Piyush Makhija, and Anuj Gupta. 2020.
\newblock \href
  {https://ui.adsabs.harvard.edu/abs/2020arXiv200312932K/abstract} {{User
  Generated Data: Achilles' Heel of BERT}}.
\newblock \emph{arXiv e-prints}, pages arXiv--2003.

\bibitem[{Lafferty et~al.(2001)Lafferty, McCallum, and
  Pereira}]{laffertyconditional}
John Lafferty, Andrew McCallum, and Fernando Pereira. 2001.
\newblock \href {https://dl.acm.org/doi/10.5555/645530.655813} {{Conditional
  Random Fields: Probabilistic Models for Segmenting and Labeling Sequence
  Data}}.
\newblock \emph{Association for Computing Machinery (ACM)}.

\bibitem[{Lample et~al.(2016)Lample, Ballesteros, Subramanian, Kawakami, and
  Dyer}]{lample2016neural}
Guillaume Lample, Miguel Ballesteros, Sandeep Subramanian, Kazuya Kawakami, and
  Chris Dyer. 2016.
\newblock \href
  {http://m-mitchell.com/NAACL-2016/NAACL-HLT2016/pdf/N16-1030.pdf} {Neural
  architectures for named entity recognition}.
\newblock In \emph{Proceedings of NAACL-HLT}, pages 260--270.

\bibitem[{Lee et~al.(2019)Lee, Yoon, Kim, Kim, Kim, So, and
  Kang}]{lee_biobert_2019}
Jinhyuk Lee, Wonjin Yoon, Sungdong Kim, Donghyeon Kim, Sunkyu Kim, Chan~Ho So,
  and Jaewoo Kang. 2019.
\newblock \href {https://doi.org/10.1093/bioinformatics/btz682} {{BioBERT}: a
  pre-trained biomedical language representation model for biomedical text
  mining}.
\newblock \emph{Bioinformatics}, page btz682.
\newblock ArXiv: 1901.08746.

\bibitem[{Li et~al.(2020)Li, Soltanolkotabi, and Oymak}]{li2020gradient}
Mingchen Li, Mahdi Soltanolkotabi, and Samet Oymak. 2020.
\newblock \href {http://proceedings.mlr.press/v108/li20j.html} {Gradient
  descent with early stopping is provably robust to label noise for
  overparameterized neural networks}.
\newblock In \emph{International Conference on Artificial Intelligence and
  Statistics}, pages 4313--4324. PMLR.

\bibitem[{Liu et~al.(2020)Liu, Niles-Weed, Razavian, and
  Fernandez-Granda}]{liu2020early}
Sheng Liu, Jonathan Niles-Weed, Narges Razavian, and Carlos Fernandez-Granda.
  2020.
\newblock \href
  {https://proceedings.neurips.cc/paper/2020/hash/ea89621bee7c88b2c5be6681c8ef4906-Abstract.html}
  {Early-learning regularization prevents memorization of noisy labels}.
\newblock \emph{Advances in Neural Information Processing Systems}, 33.

\bibitem[{Liu et~al.(2019)Liu, Ott, Goyal, Du, Joshi, Chen, Levy, Lewis,
  Zettlemoyer, and Stoyanov}]{liu_roberta_2019}
Yinhan Liu, Myle Ott, Naman Goyal, Jingfei Du, Mandar Joshi, Danqi Chen, Omer
  Levy, Mike Lewis, Luke Zettlemoyer, and Veselin Stoyanov. 2019.
\newblock \href {http://arxiv.org/abs/1907.11692} {{RoBERTa}: {A} {Robustly}
  {Optimized} {BERT} {Pretraining} {Approach}}.
\newblock \emph{arXiv:1907.11692 [cs]}.
\newblock ArXiv: 1907.11692.

\bibitem[{Loshchilov and Hutter(2019)}]{loshchilov_decoupled_2019}
Ilya Loshchilov and Frank Hutter. 2019.
\newblock \href {http://arxiv.org/abs/1711.05101} {Decoupled {Weight} {Decay}
  {Regularization}}.
\newblock \emph{arXiv:1711.05101 [cs, math]}.
\newblock ArXiv: 1711.05101.

\bibitem[{Peters et~al.(2018)Peters, Neumann, Iyyer, Gardner, Clark, Lee, and
  Zettlemoyer}]{Peters2018elmo}
Matthew~E. Peters, Mark Neumann, Mohit Iyyer, Matt Gardner, Christopher Clark,
  Kenton Lee, and Luke Zettlemoyer. 2018.
\newblock \href {http://arxiv.org/abs/arXiv:1802.05365v1} {{Deep contextualized
  word representations}}.
\newblock In \emph{Proceedings of NAACL-HLT 2018}.

\bibitem[{Petroni et~al.(2019)Petroni, Rockt{\"{a}}schel, Lewis, Bakhtin, Wu,
  Miller, and Riedel}]{Petroni2019knowledge_bases}
Fabio Petroni, Tim Rockt{\"{a}}schel, Patrick Lewis, Anton Bakhtin, Yuxiang Wu,
  Alexander~H. Miller, and Sebastian Riedel. 2019.
\newblock \href {http://arxiv.org/abs/1909.01066} {{Language Models as
  Knowledge Bases?}}
\newblock In \emph{Proceedings of EMNLP 2019}.

\bibitem[{Rogers et~al.(2020)Rogers, Kovaleva, and Rumshisky}]{Rogers2020}
Anna Rogers, Olga Kovaleva, and Anna Rumshisky. 2020.
\newblock \href {https://doi.org/10.1162/tacl_a_00349} {A primer in
  {BERT}ology: What we know about how {BERT} works}.
\newblock \emph{Transactions of the Association for Computational Linguistics},
  8:842--866.

\bibitem[{Sang and De~Meulder(2003)}]{sang_introduction_2003}
Erik F. Tjong~Kim Sang and Fien De~Meulder. 2003.
\newblock \href {http://arxiv.org/abs/cs/0306050} {Introduction to the
  {CoNLL}-2003 {Shared} {Task}: {Language}-{Independent} {Named} {Entity}
  {Recognition}}.
\newblock \emph{arXiv:cs/0306050}.
\newblock ArXiv: cs/0306050.

\bibitem[{Snell et~al.(2017)Snell, Swersky, and
  Zemel}]{snell_prototypical_2017}
Jake Snell, Kevin Swersky, and Richard~S. Zemel. 2017.
\newblock \href {http://arxiv.org/abs/1703.05175} {Prototypical {Networks} for
  {Few}-shot {Learning}}.
\newblock \emph{arXiv:1703.05175 [cs, stat]}.
\newblock ArXiv: 1703.05175.

\bibitem[{Strubell et~al.(2019)Strubell, Ganesh, and
  McCallum}]{strubell-etal-2019-energy}
Emma Strubell, Ananya Ganesh, and Andrew McCallum. 2019.
\newblock \href {https://doi.org/10.18653/v1/P19-1355} {Energy and policy
  considerations for deep learning in {NLP}}.
\newblock In \emph{Proceedings of the 57th Annual Meeting of the Association
  for Computational Linguistics}, pages 3645--3650, Florence, Italy.
  Association for Computational Linguistics.

\bibitem[{Tenney et~al.(2019)Tenney, Das, and Pavlick}]{tenney2019bert}
Ian Tenney, Dipanjan Das, and Ellie Pavlick. 2019.
\newblock \href {https://aclanthology.org/P19-1452} {{BERT Rediscovers the
  Classical NLP Pipeline}}.
\newblock In \emph{Proceedings of the 57th Annual Meeting of the Association
  for Computational Linguistics}, pages 4593--4601.

\bibitem[{Toneva et~al.(2019)Toneva, Sordoni, {Tachet des Combes}, Trischler,
  Bengio, and Gordon}]{toneva_empirical_2019}
Mariya Toneva, Alessandro Sordoni, Remi {Tachet des Combes}, Adam Trischler,
  Yoshua Bengio, and Geoffrey~J. Gordon. 2019.
\newblock \href {https://arxiv.org/abs/1812.05159} {{An Empirical Study of
  Example Forgetting during Deep Neural Network Learning}}.
\newblock In \emph{Proceedings of ICLR 2019}.

\bibitem[{Tu et~al.(2020)Tu, Lalwani, Gella, and He}]{tu-etal-2020-empirical}
Lifu Tu, Garima Lalwani, Spandana Gella, and He~He. 2020.
\newblock \href {https://doi.org/10.1162/tacl_a_00335} {An empirical study on
  robustness to spurious correlations using pre-trained language models}.
\newblock \emph{Transactions of the Association for Computational Linguistics},
  8:621--633.

\bibitem[{Wang et~al.(2019)Wang, Shang, Liu, Lu, Liu, and
  Han}]{wang_crossweigh_2019}
Zihan Wang, Jingbo Shang, Liyuan Liu, Lihao Lu, Jiacheng Liu, and Jiawei Han.
  2019.
\newblock \href {http://arxiv.org/abs/1909.01441} {{CrossWeigh}: {Training}
  {Named} {Entity} {Tagger} from {Imperfect} {Annotations}}.
\newblock \emph{arXiv:1909.01441 [cs]}.
\newblock ArXiv: 1909.01441.

\bibitem[{Xie et~al.(2017)Xie, Girshick, Dollár, Tu, and
  He}]{xie_aggregated_2017}
Saining Xie, Ross Girshick, Piotr Dollár, Zhuowen Tu, and Kaiming He. 2017.
\newblock \href {http://arxiv.org/abs/1611.05431} {Aggregated {Residual}
  {Transformations} for {Deep} {Neural} {Networks}}.
\newblock \emph{arXiv:1611.05431 [cs]}.
\newblock ArXiv: 1611.05431.

\bibitem[{Zhang et~al.(2017)Zhang, Bengio, Hardt, Recht, and
  Vinyals}]{Zhang2017understanding_deep_learning}
Chiyuan Zhang, Samy Bengio, Moritz Hardt, Benjamin Recht, and Oriol Vinyals.
  2017.
\newblock \href {https://dl.acm.org/doi/abs/10.1145/3446776} {{Understanding
  deep learning requires rethinking generalization}}.
\newblock In \emph{Proceedings of ICLR 2017}.

\bibitem[{Zhang et~al.(2021)Zhang, Bengio, Hardt, Recht, and
  Vinyals}]{zhang2021understanding}
Chiyuan Zhang, Samy Bengio, Moritz Hardt, Benjamin Recht, and Oriol Vinyals.
  2021.
\newblock \href {https://doi.org/10.1145/3446776} {Understanding deep learning
  (still) requires rethinking generalization}.
\newblock \emph{Commun. ACM}, 64(3):107–115.

\end{thebibliography}
\bibliographystyle{acl_natbib}

\clearpage

\appendix

\section{Comparison of learning phases in a BiLSTM and ResNet on CIFAR-10} \label{app:second_phase_comparison}

\begin{figure}[hbt]
    \centering
	\footnotesize
    \includegraphics[width=0.95\linewidth]{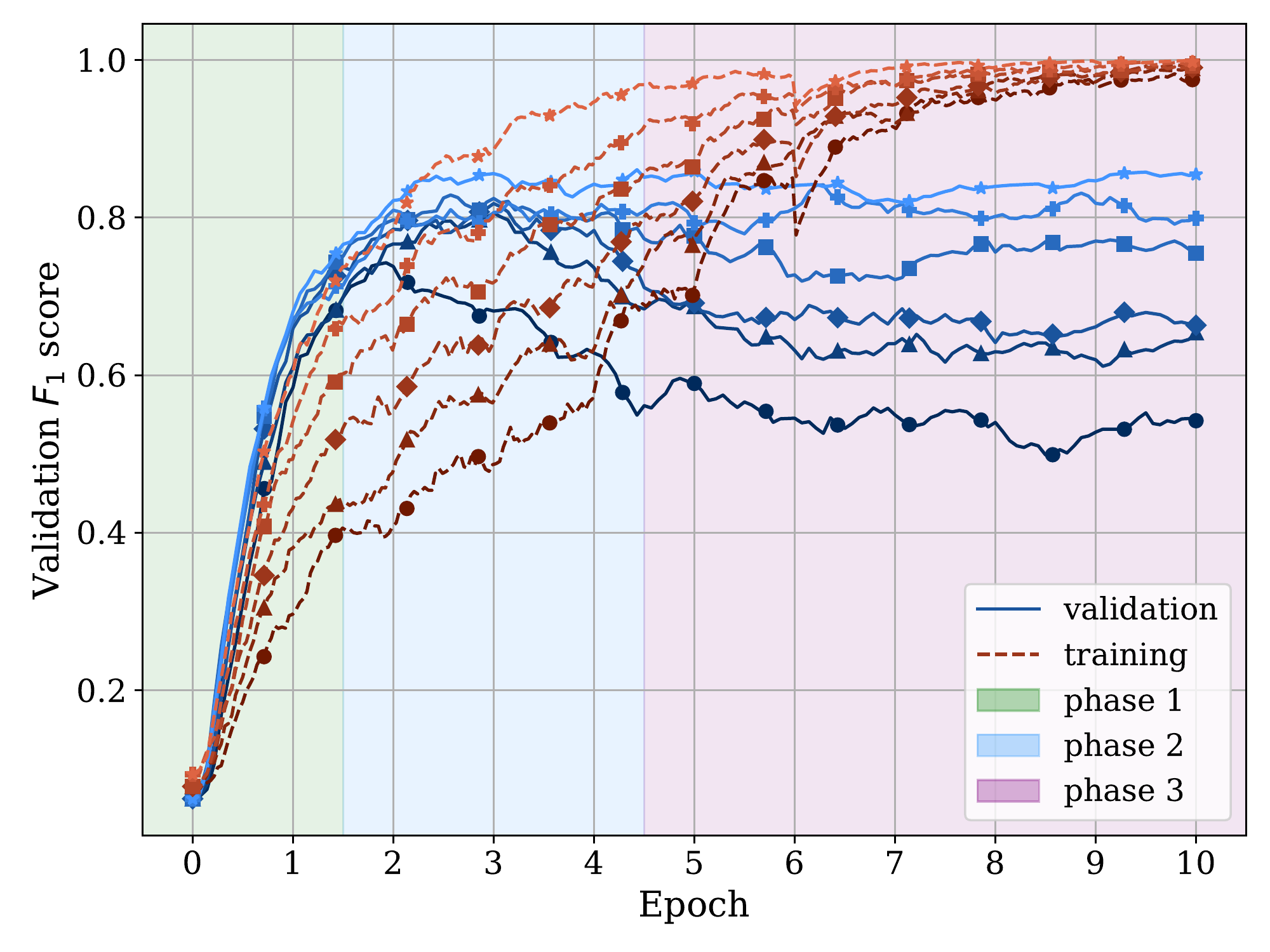}
    \caption{Performance (F$_1$) of a ResNet model throughout the training process on the \texttt{CIFAR10} dataset. Darker colours correspond to higher levels of noise (0\% to 50\%).}
    \label{fig:noise-performance-specific-resnet}
\end{figure}

For comparison, we show the training progress of a ResNet \cite{he_deep_2015} trained on \texttt{CIFAR10} \cite{krizhevsky_learning_2009} in Figure \ref{fig:noise-performance-specific-resnet}. Following \citet{toneva_empirical_2019}, we use a ResNeXt model \cite{xie_aggregated_2017} with 101 blocks pre-trained on the ImageNet dataset \cite{deng2009imagenet}. The model has been fine-tuned with a cross-entropy loss with the same optimiser and hyper-parameters as BERT. We evaluate it using F$_1$ score. As can be seen, the training performance continues to increase while the validation performs plateaus or decreases, with no clearly delineated second phase as in the pre-trained BERT's training.

\section{JNLPBA noise results} \label{app:jnlpba_results}

\begin{figure}[hbt]
	\centering
	\footnotesize
	\includegraphics[width=0.95\linewidth]{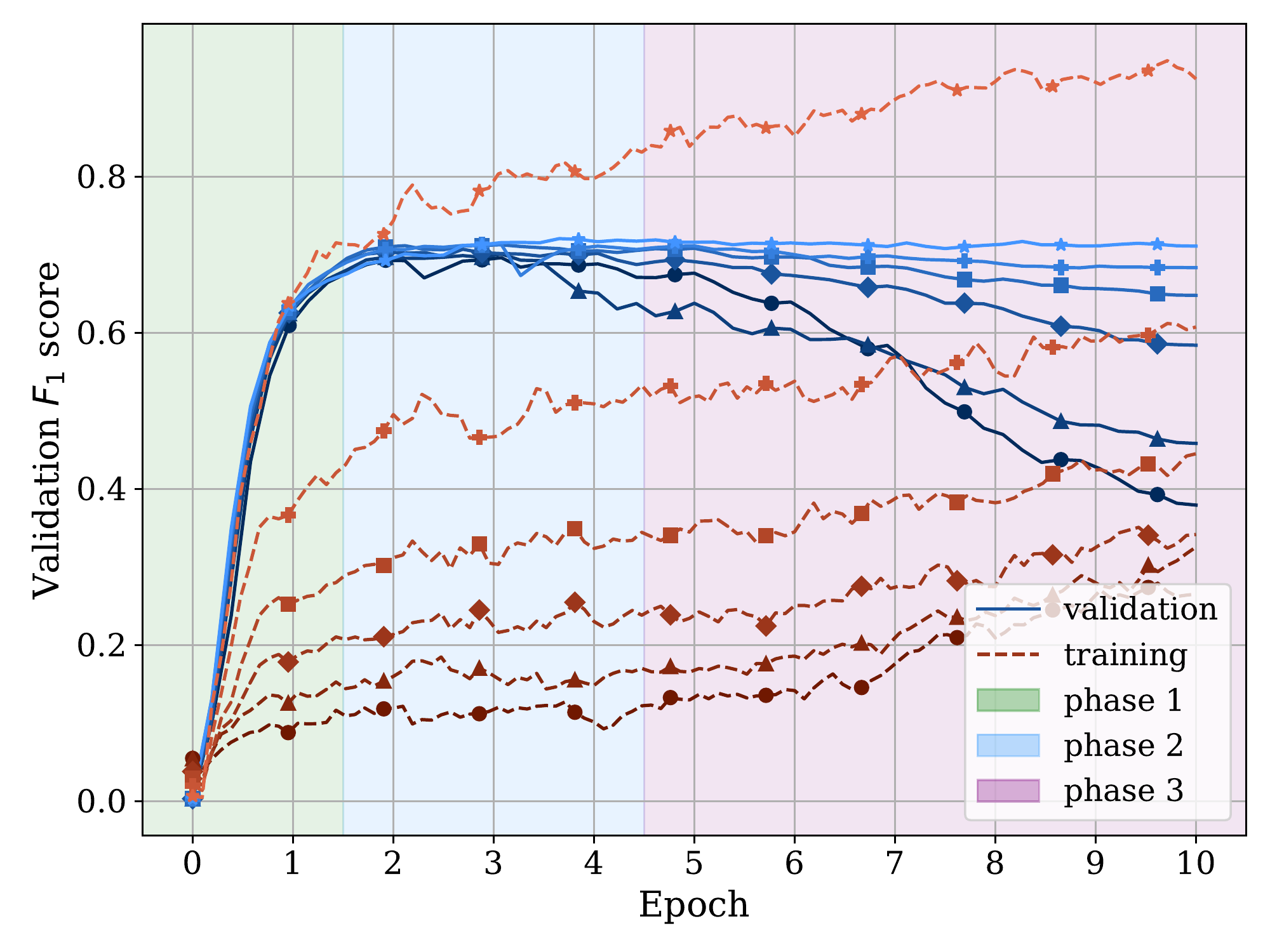}
    \caption{BERT performance (F$_1$) throughout the training process on the \texttt{JNLPBA} dataset. Darker colours correspond to higher levels of noise (0\% to 50\%).}
	\label{fig:noise-performance-specific-paper-jnlpba}
\end{figure}

\begin{figure}[hbt]
	\centering
	\footnotesize
	\includegraphics[width=0.95\linewidth]{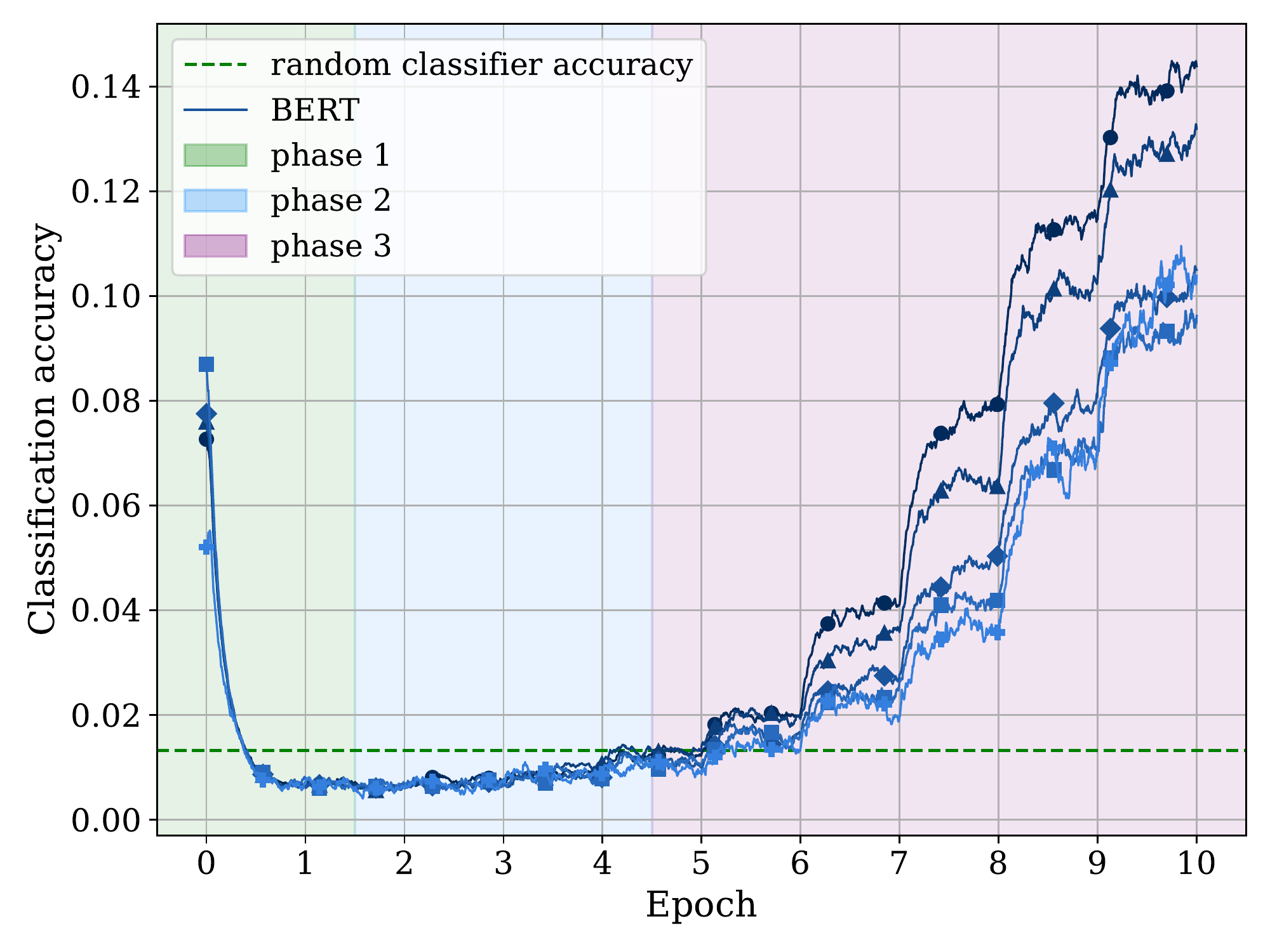}
    \caption{Classification accuracy of noisy examples in the training set for the \texttt{JNLPBA} dataset. Darker colours correspond to higher levels of noise (0\% to 50\%).}
	\label{fig:accuracy-noisy-paper-jnlpba}
\end{figure}

As well as \texttt{CoNLL03}, we also report the analysis on the \texttt{JNLPBA} dataset. In Figure \ref{fig:noise-performance-specific-paper-jnlpba}, we show the performance of BERT on increasingly noisy versions of the training set. In Figure \ref{fig:accuracy-noisy-paper-jnlpba}, we report the accuracy of noisy examples.

\section{Effect of pre-training} \label{app:effect_of_pretraining}

\begin{figure}[bt]
	\footnotesize
	\centering
	\includegraphics[width=0.95\linewidth]{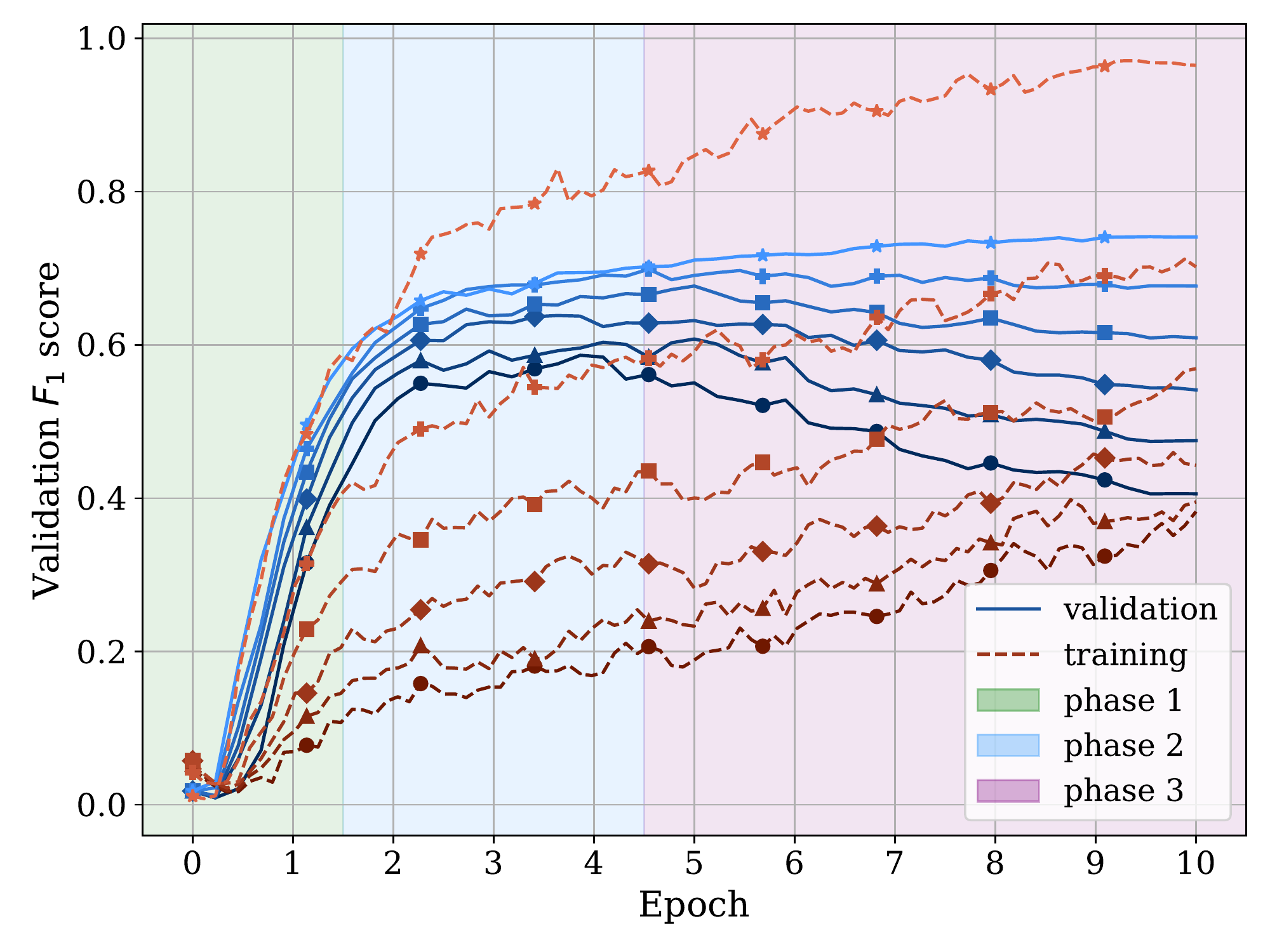}
    \caption{Performance (F$_1$) of a non-pre-trained BERT model throughout the training process on the \texttt{CoNLL03} train and validation sets. Darker colours correspond to higher levels of noise (0\% to 50\%).}
	\label{fig:noise-performance-specific-paper-nopret}
\end{figure}

\begin{figure}[bt]
	\footnotesize
	\centering
	\includegraphics[width=0.95\linewidth]{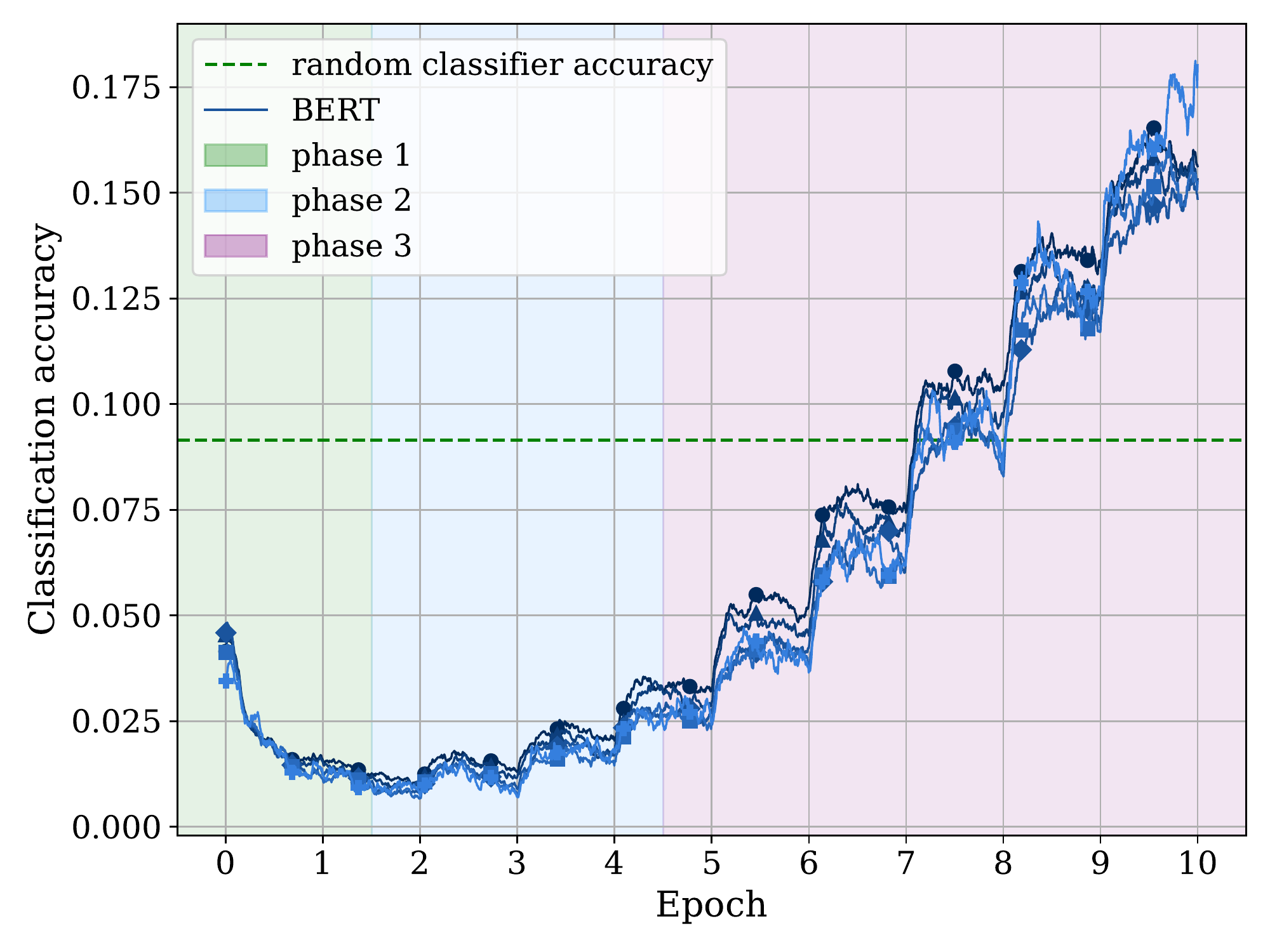}
    \caption{Classification accuracy of a non-pre-trained BERT model on noisy examples in the training set for the \texttt{CoNLL03} dataset. Darker colours correspond to higher levels of noise (0\% to 50\%).}
	\label{fig:accuracy-noisy-paper-nopret}
\end{figure}

BERT's second phase of pre-training and noise resilience are mainly attributable to its pre-training. We show the training progress of a non-pretrained BERT model on \texttt{CoNLL03} in Figure \ref{fig:noise-performance-specific-paper-nopret} and its classification accuracy on noisy examples in Figure \ref{fig:accuracy-noisy-paper-nopret}. As can be seen, a non-pre-trained BERT's training performance continuously improves and so does its performance on noisy examples.

\section{Examples of forgettable examples} \label{sec:forgettable_examples}
In Table \ref{tab:examples-forg-conll}, we can find the sentences containing the most forgettable examples during a training run of 50 epochs for the \texttt{CoNLL03} dataset. The maximum theoretical number of forgetting events in this case is 25. It is important to notice how the most forgotten entity presents a mismatched "The", which the network correctly classifies as an "other" (\texttt{O}) entity.

\begin{table*}[bt]
\centering
\begin{tabular}{@{}l|r@{}}
\toprule
\textbf{Sentence}                                                                              & \textbf{Number of forgetting events} \\ \midrule
the third and final test between England and Pakistan at \textbf{The} (I-LOC) & 11                                   \\
\textbf{GOLF} - BRITISH MASTERS THIRD ROUND SCORES . (O)                      & 10                                   \\
\textbf{GOLF} - GERMAN OPEN FIRST ROUND SCORES . (O)                          & 10                                   \\
\textbf{English County Championship} cricket matches on Saturday : (MISC)     & 10                                   \\
\textbf{English County Championship} cricket matches on Friday : (MISC)       & 9                                    \\ \bottomrule
\end{tabular}
\caption{Sentences containing the most forgettable examples in the \texttt{CoNLL03} dataset. In bold the entity that was most often forgotten within the given sentence and in brackets its ground-truth classification.}
\label{tab:examples-forg-conll}
\end{table*}

\section{BERT as a noise detector}\label{app:noise-detection}
We report the exact detection metrics for the model proposed in section \ref{sec:noise-effect} in Table \ref{tab:noise-complicated}. Here we can see how both for extremely noisy datasets and for cleaner datasets, our model is able to detect the noisy examples with about 90-91\% F$_1$ score, as mentioned above. 

\begin{table}[hbt]
	\centering
	\footnotesize
	\begin{tabular}{r|rrr} 
		\toprule
		\textbf{Noise} & \textbf{Precision} & \textbf{Recall} & \textbf{F$_1$ score}\\ 
		\midrule
		10\%                       & 92.18\%                        & 95.90\%                     & 94.00\%                       \\
		20\%                       & 96.19\%                        & 96.33\%                     & 96.26\%                       \\
		30\%                       & 98.02\%                        & 96.35\%                     & 97.17\%                       \\
		40\%                       & 98.27\%                        & 96.95\%                     & 97.60\%                       \\
		50\%                       & 98.64\%                        & 97.27\%                     & 97.94\%                       \\
		\bottomrule
	\end{tabular}
	\caption{Noise detection performance with varying levels of noise on the \texttt{CoNLL03} dataset using the method proposed.}
	\label{tab:noise-complicated}
\end{table}

Moreover, we provide the implementation used to detect outliers used to produce the table and figures above:
\begin{enumerate}
	\item We first collect the losses for each training example after a short fine-tuning process (4 epochs in our case).
	\item We then assume an unknown portion of these examples is noisy, giving rise to a two-class classification problem (noisy vs non-noisy). To discriminate the two classes, we then solve the following optimisation problem which aims to find a loss threshold $T$ that minimises inter-class variance for each of the two classes:
	\begin{equation*}
		\underset{ T }{\arg \min } \sum_{ x\ <\ T}\left\| x - \mu _{c}\right\|^{2} + \sum_{ x\ \geq\ T}\left\| x - \mu _{n}\right\|^{2}
	\end{equation*}
	Where elements denoted as $x$ are the losses extracted from the training set, $\mu_c$ is the mean of all $x < T$, and $\mu_n$ is the mean of all $x \geq T$.
	\item For testing purposes, we then apply the method to the chosen training set and measure the noise detection F1 score.
\end{enumerate}

\begin{figure}[hbt]
	\centering
	\includegraphics[width=\linewidth]{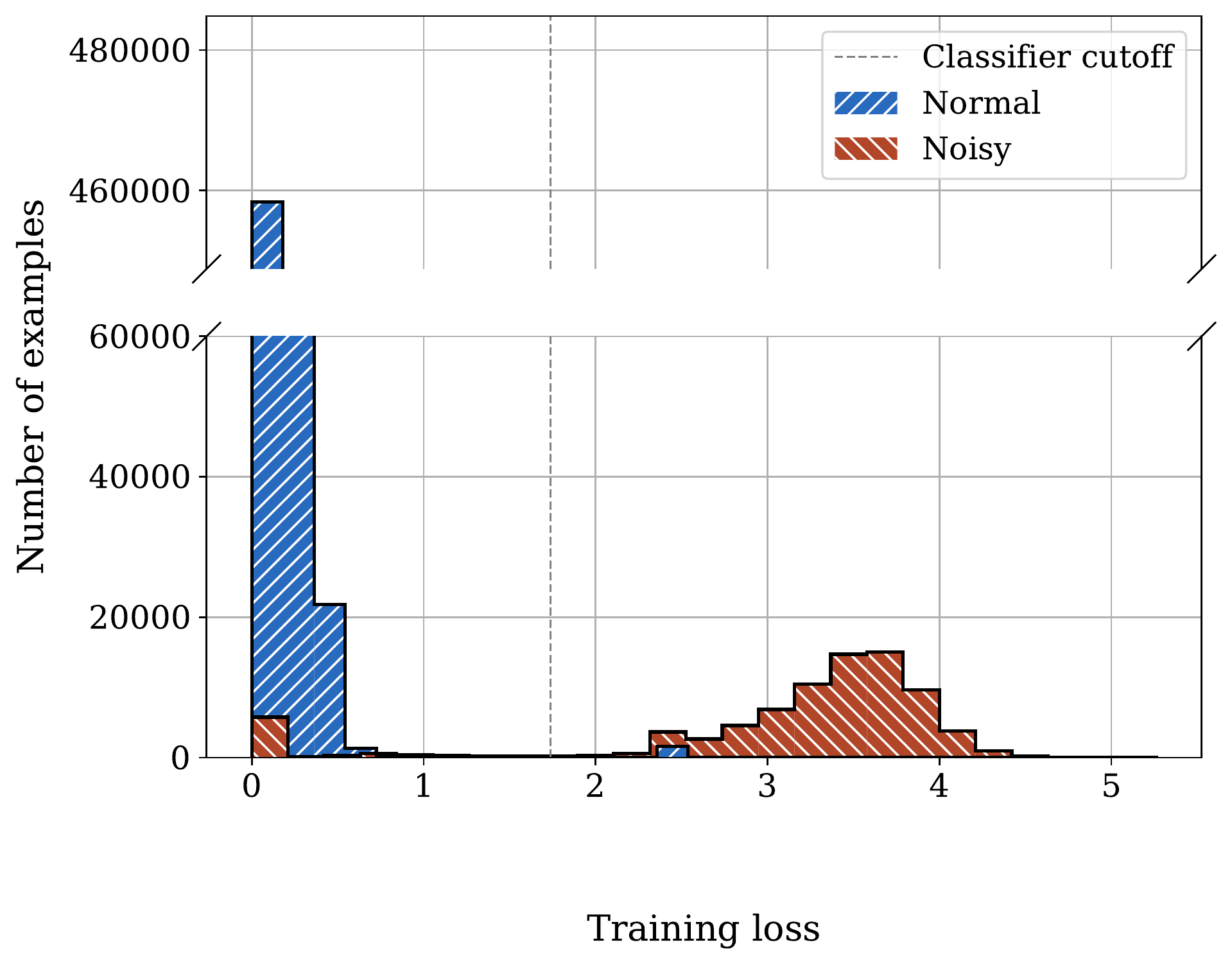}
	\caption{Loss distribution for noisy and non-noisy examples from the \texttt{CoNLL03} training set. The grey dashed line represent the chosen loss threshold found by our method to discriminate between noisy and non-noisy examples.}
	\label{fig:histogram_losses}
\end{figure}

In Figure \ref{fig:histogram_losses}, we qualitatively saw how the losses are distributed for noisy and regular examples and notice how they are neatly separated except for a small subset of the noisy examples. These examples might have been already memorised by the model, which would explain their lower loss.

\section{JNLPBA forgetting results} \label{app:jnlpba_forgetting}

\begin{figure}[hbt]
	\centering
	\footnotesize
	\includegraphics[width=0.95\linewidth]{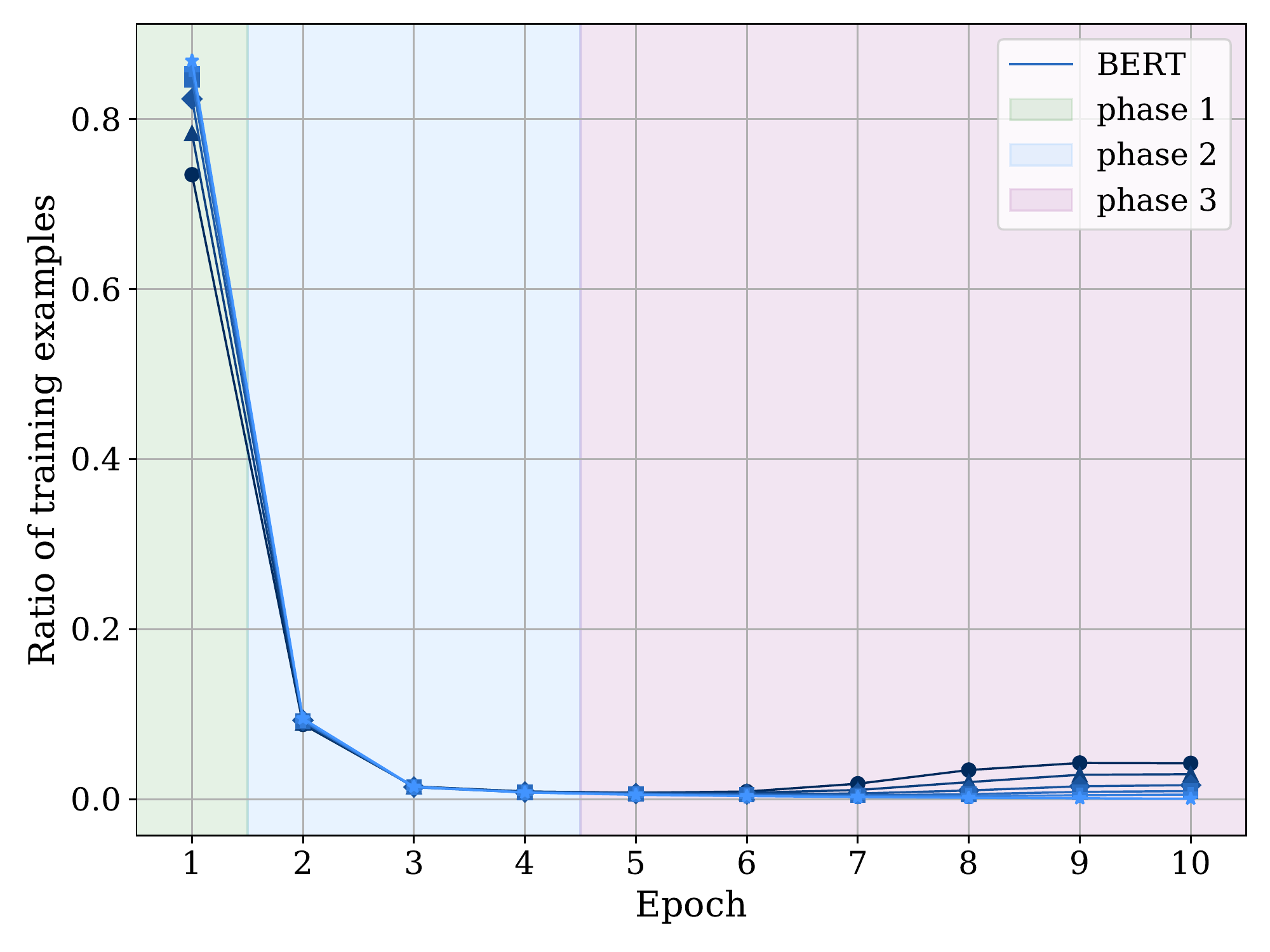}
	\caption{First learning events distribution during BERT training for various levels of noise on the \texttt{JNLPBA} dataset. Darker colours correspond to higher levels of noise (0\% to 50\%).}
	\label{fig:histo-fle-paper-jnlpba}
\end{figure}

We show in Figure \ref{fig:histo-fle-paper-jnlpba} how many data points were learned by BERT for the first time at each epoch on the \texttt{JNLPBA} dataset during training (first learning events).

\section{Further ProtoBERT results}\label{app:further-results}
As in Table \ref{tab:base-vs-proto} we only reported F$_1$ score for our methods, for completeness we also report precision and recall in table \ref{tab:base-vs-proto-complete}.

\begin{table*}[tb]
	\footnotesize
	\centering
	\begin{tabular}{@{}l|crr|crr|crr@{}}
		\toprule
		\multicolumn{1}{l|}{\multirow{2}{*}{\textbf{Model}}} & \multicolumn{3}{c|}{\textbf{CoNLL03}}                                                                     & \multicolumn{3}{c|}{\textbf{JNLPBA}}                                                                      & \multicolumn{3}{c}{\textbf{WNUT17}}                                                                     \\ \cmidrule(l){2-10} 
		\multicolumn{1}{c|}{}                                & \textbf{P}                         & \multicolumn{1}{c}{\textbf{R}} & \multicolumn{1}{c|}{\textbf{F$_1$}} & \textbf{P}                         & \multicolumn{1}{c}{\textbf{R}} & \multicolumn{1}{c|}{\textbf{F$_1$}} & \textbf{P}                         & \multicolumn{1}{c}{\textbf{R}} & \multicolumn{1}{c}{\textbf{F$_1$}} \\ \midrule
		State-of-the-art                  & NA & NA                 & 93.50                     & NA          & NA                 & 77.59                               & NA & NA                 & 50.03                  \\
		BERT + classification layer (baseline)                          & 88.97          & 89.75                          & 89.35                               & \textbf{72.99} & 77.90                          & \textbf{75.36}                      & 53.65          & 37.42                          & 44.09                              \\
		\midrule
		ProtoBERT                                      & \textbf{89.26} & \textbf{90.49}                 & \textbf{89.87}                      & 68.66          & \textbf{80.03}                 & 73.91                               & \textbf{54.38} & 43.96                 & \textbf{48.62}                     \\
		ProtoBERT + running centroids                  & 89.03 & 89.91                 & 89.46                     & 68.92          & 78.83                 & 73.54                               & 54.11 & \textbf{44.05}                 & 48.56                  \\
		\bottomrule
	\end{tabular}
	\caption{Comparison between the baseline model and the proposed architecture on the \texttt{CoNLL03}, \texttt{JNLPBA} and \texttt{WNUT17} datasets evaluated using entity-level metrics.}
	\label{tab:base-vs-proto-complete}
\end{table*}

\begin{table*}[hbt]
	\centering
	\footnotesize
	\begin{tabular}{r|rrr|r} 
		\toprule
		\multicolumn{1}{l|}{\textbf{Noise}} & \multicolumn{1}{l}{\textbf{Forgettable}} & \multicolumn{1}{l}{\textbf{Unforgettable}} & \multicolumn{1}{l}{\textbf{Learned}} & \multicolumn{1}{|l}{\textbf{Forgettable/learned (\%)}}  \\ 
		\midrule
		\texttt{CoNLL03} 0\%                        & 2,669                           & 699,381                            & 230,716                      & 1.1568\%                                     \\
		\texttt{CoNLL03} 10\%                       & 10,352                          & 691,698                            & 224,968                      & 4.6015\%                                     \\
		\texttt{CoNLL03} 20\%                       & 19,667                          & 682,383                            & 216,780                      & 9.0723\%                                     \\
		\texttt{CoNLL03} 30\%                       & 30,041                          & 672,009                            & 209,191                      & 14.3606\%                                    \\
		\midrule
		\texttt{JNLPBA} 0\%                        & 23,263                           & 817,087                            & 457,485                      & 5.0849\%                                     \\
		\texttt{JNLPBA} 10\%                       & 26,667                          & 813,683                            & 422,264                      & 6.3152\%                                     \\
		\texttt{JNLPBA} 20\%                       & 26,369                          & 813,981                            & 386,562                      & 6.8214\%                                     \\
		\texttt{JNLPBA} 30\%                       & 30,183                          & 810,167                            & 353,058 & 8.5490\%\\
		\midrule
		\texttt{CIFAR10} 0\%                        & 8,328                           & 36,672                             & 45,000                       & 18.5067\%                                    \\
		\texttt{CIFAR10} 10\%                       & 9,566                           & 35,434                             & 44,976                       & 21.2691\%                                    \\
		\texttt{CIFAR10} 20\%                       & 9,663                           & 35,337                             & 44,922                       & 21.5106\%                                    \\
		\texttt{CIFAR10} 30\%                       & 11,207                          & 33,793                             & 44,922                       & 24.9477\% \\        
		\bottomrule
	\end{tabular}
	\caption{Number of forgettable, unforgettable, and learned examples during BERT training on the \texttt{CoNLL03}, \texttt{JNLPBA} and \texttt{CIFAR10} datasets.}
	\label{tab:forgetting-events-bert-resnet}
\end{table*}

\begin{table}[hbt!]
	\centering
	\footnotesize
    \begin{tabular}{r|rr}
    	\toprule
    	\textbf{Examples} & \textbf{BERT} & \textbf{bi-LSTM} \\
    	\midrule
    	Forgettable & 2,669 & 144,377 \\
    	Unforgettable & 699,381 & 60,190 \\
    	Learned & 230,716 & 184,716 \\
		\midrule
    	Forgettable/learned (\%) & 1.1568\% & 78,1616\% \\
    	\bottomrule
    \end{tabular}
	\caption{Comparison of the number of forgettable, learnable and unforgettable examples between BERT and a bi-LSTM model.}
\label{tab:comparison-BERT-LSTM}
\end{table}

\section{ProtoBERT results on JNLPBA}\label{app:proto-jnlpba}

\begin{figure}[hbt]
	\centering
	\footnotesize
	\includegraphics[width=0.95\linewidth]{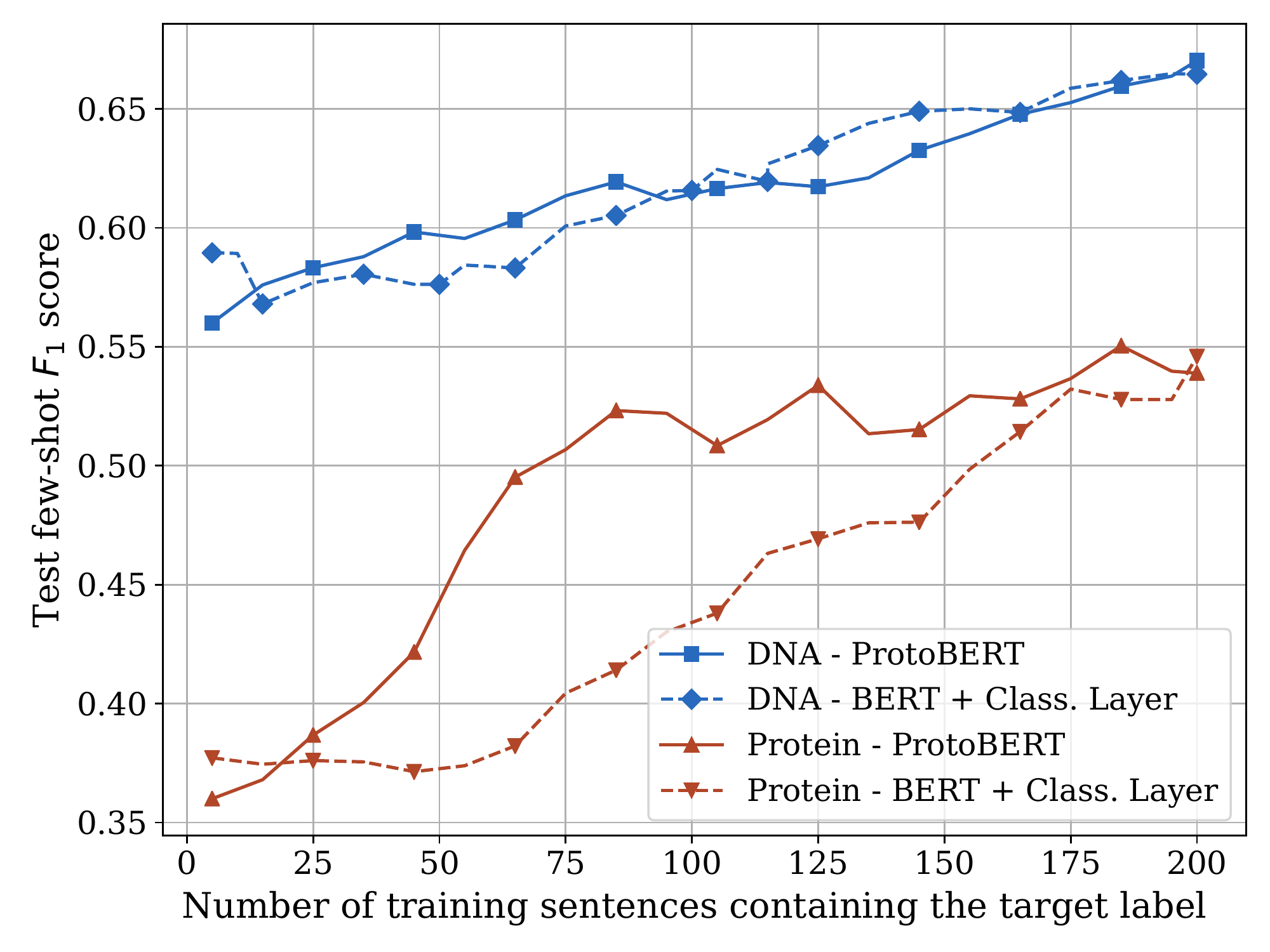}
	\caption{Model performance comparison between the baseline model and ProtoBERT for the \texttt{JNLPBA} dataset, reducing the sentences containing the \texttt{DNA} and \texttt{Protein} class. Results reported as F$_1$ score on all classes.}
	\label{fig:jnlpba-proto-full-paper-final-2}
\end{figure}

We report in Figure \ref{fig:jnlpba-proto-full-paper-final-2} the comparison between our baseline and ProtoBERT for all classes.

\section{Results on other pretrained transformers}\label{sec:other-transformers}
While most of the main paper focuses on BERT, it is worthwhile to mention the results on other pre-trained transformers and compare the results.

\begin{figure}[hbt]
	\centering
	\footnotesize
	\includegraphics[width=0.95\linewidth]{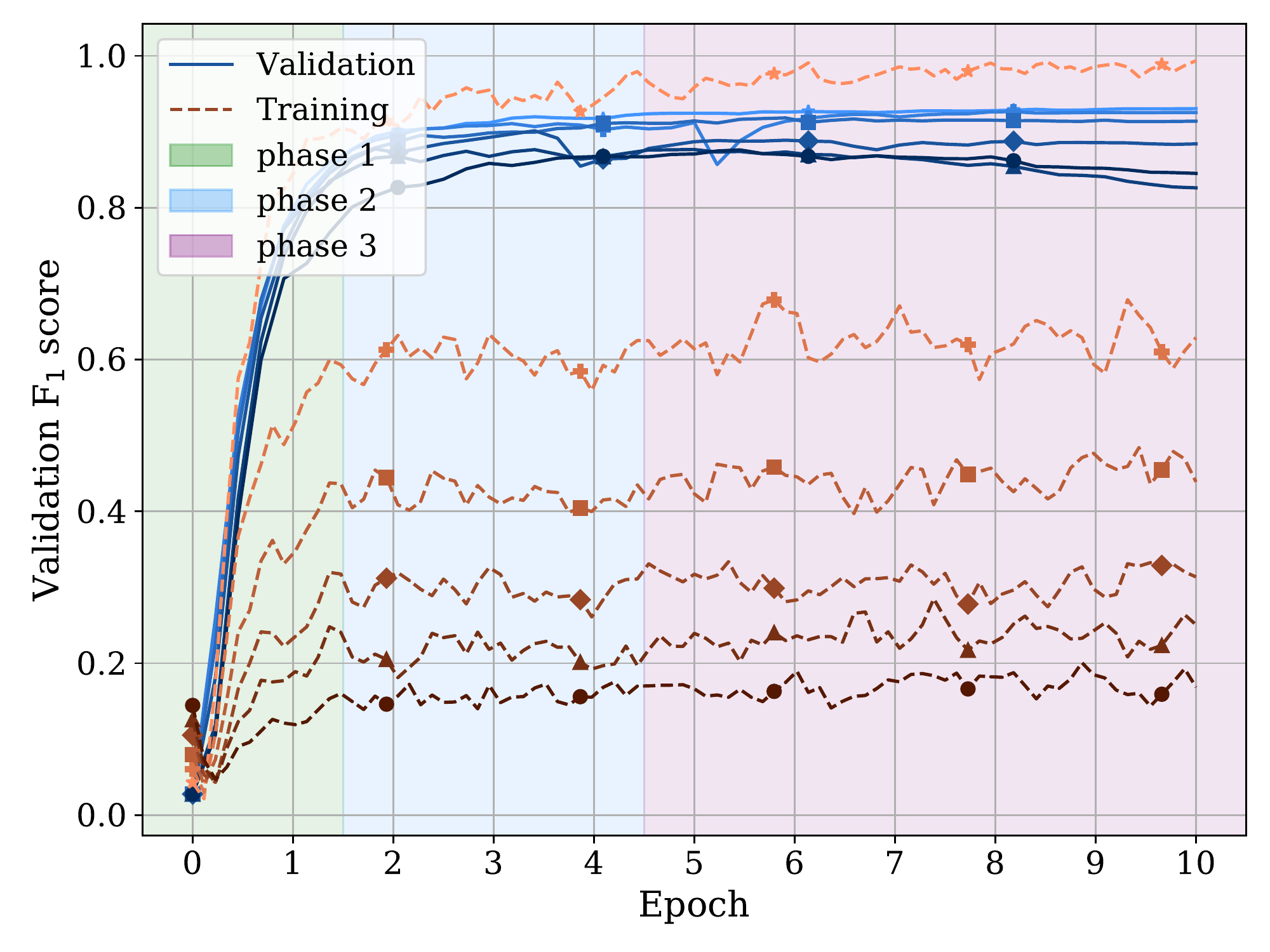}
	\caption{RoBERTa performance (F$_1$) throughout the training process on the \texttt{CoNLL03} train and validation sets. Darker colours correspond to higher levels of noise (0\% to 50\%).}
	\label{fig:roberta-f1-noisy}
\end{figure}

\begin{figure}[hbt]
	\centering
	\footnotesize
	\includegraphics[width=0.95\linewidth]{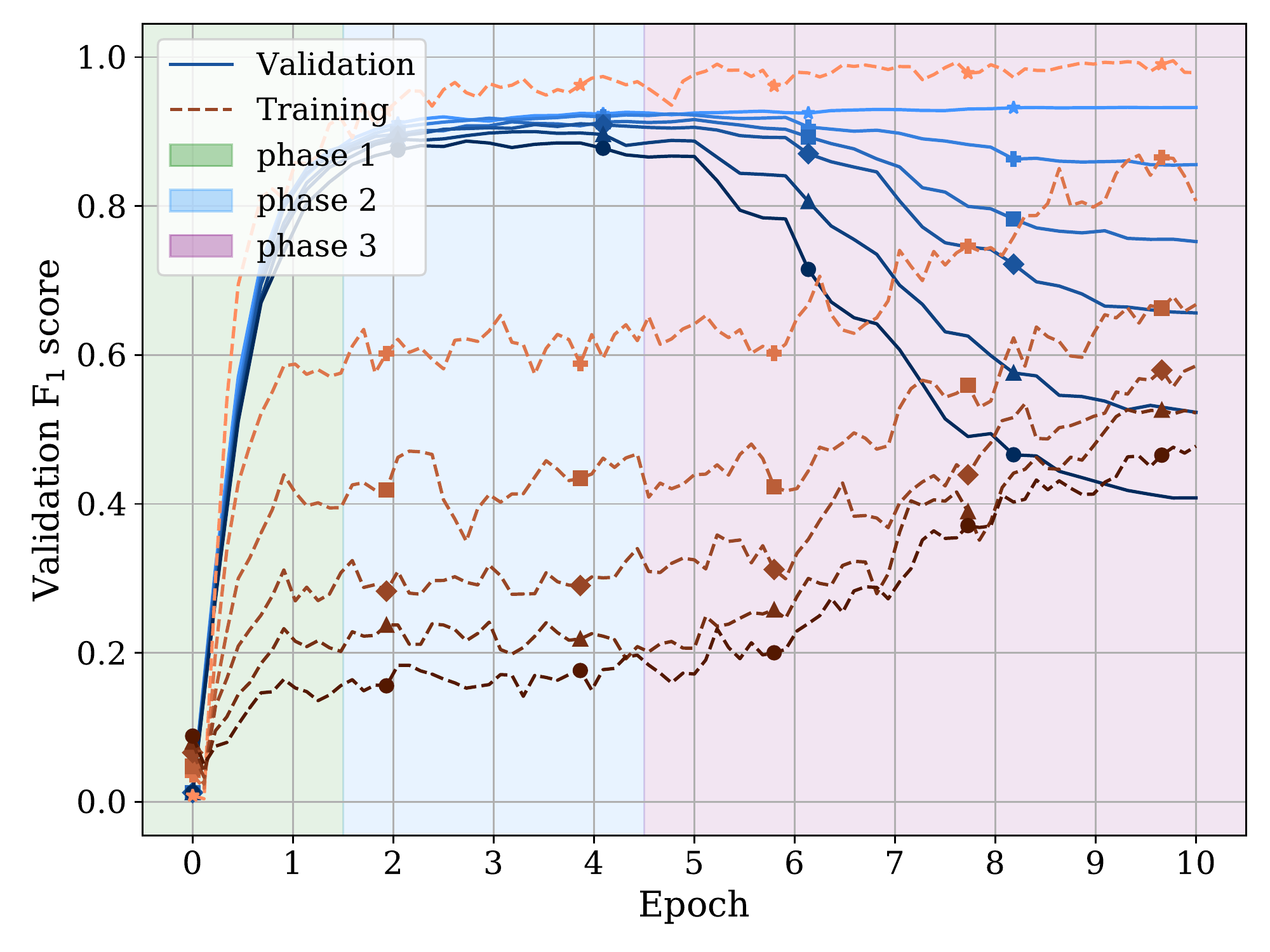}
	\caption{DeBERTa performance (F$_1$) throughout the training process on the \texttt{CoNLL03} train and validation sets. Darker colours correspond to higher levels of noise (0\% to 50\%).}
	\label{fig:deberta-f1-noisy}
\end{figure}

In Figures \ref{fig:roberta-f1-noisy} and \ref{fig:deberta-f1-noisy}, we show the validation performances (classification F$_1$ score) for the \texttt{CoNLL03} datasets for the RoBERTa and DeBERTa models (similarly to Figure \ref{fig:noise-performance-specific}). We notice that the three phases of training reported above are apparent in all studied models. RoBERTa, in particular, displays the same pattern, but shows higher robustness to noise compared to the other two models.

\begin{figure}[hbt]
	\centering
	\footnotesize
	\includegraphics[width=0.95\linewidth]{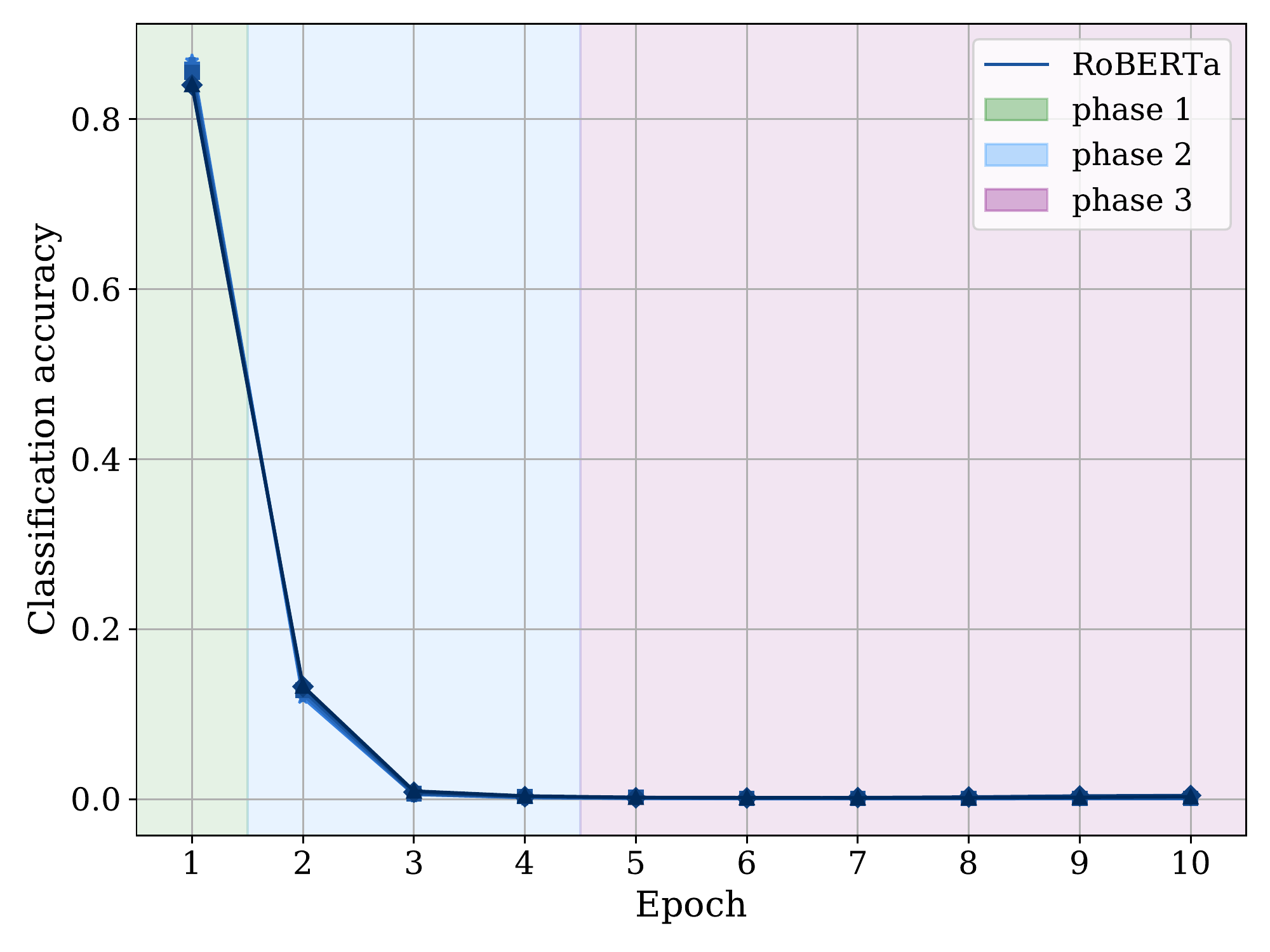}
	\caption{First learning events distribution during RoBERTa training for various levels of noise on the \texttt{CoNLL03} dataset. Darker colours correspond to higher levels of noise (0\% to 50\%).}
	\label{fig:roberta-forgetting}
\end{figure}

\begin{figure}[hbt]
	\centering
	\footnotesize
	\includegraphics[width=0.95\linewidth]{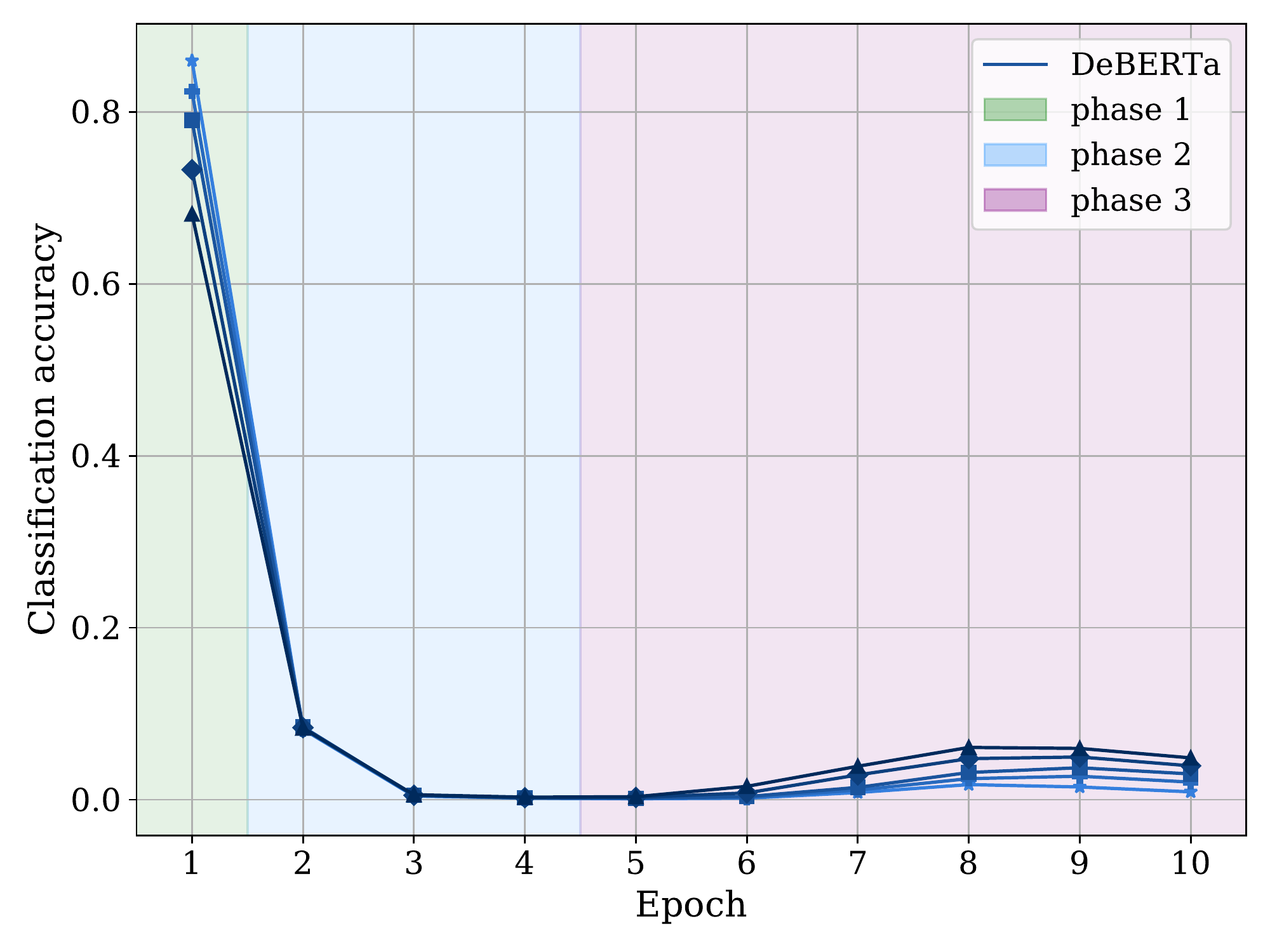}
	\caption{First learning events distribution during DeBERTa training for various levels of noise on the \texttt{CoNLL03} dataset. Darker colours correspond to higher levels of noise (0\% to 50\%).}
	\label{fig:deberta-forgetting}
\end{figure}

Moreover, in Figures \ref{fig:roberta-forgetting} and \ref{fig:deberta-forgetting}, we report the distribution of first learning events (similarly to Figure \ref{fig:fles-loc-03}) on RoBERTa and DeBERTa. As above, we can observe the same pattern described in the main body of the paper, with the notable exception that RoBERTa  is again more robust to learning the noise in later phases of the training.

\section{Few-shot \texttt{MISC} memorisation}\label{app:misc-experiments}
\begin{figure}[hbt]
	\centering
	\footnotesize
	\includegraphics[width=0.95\linewidth]{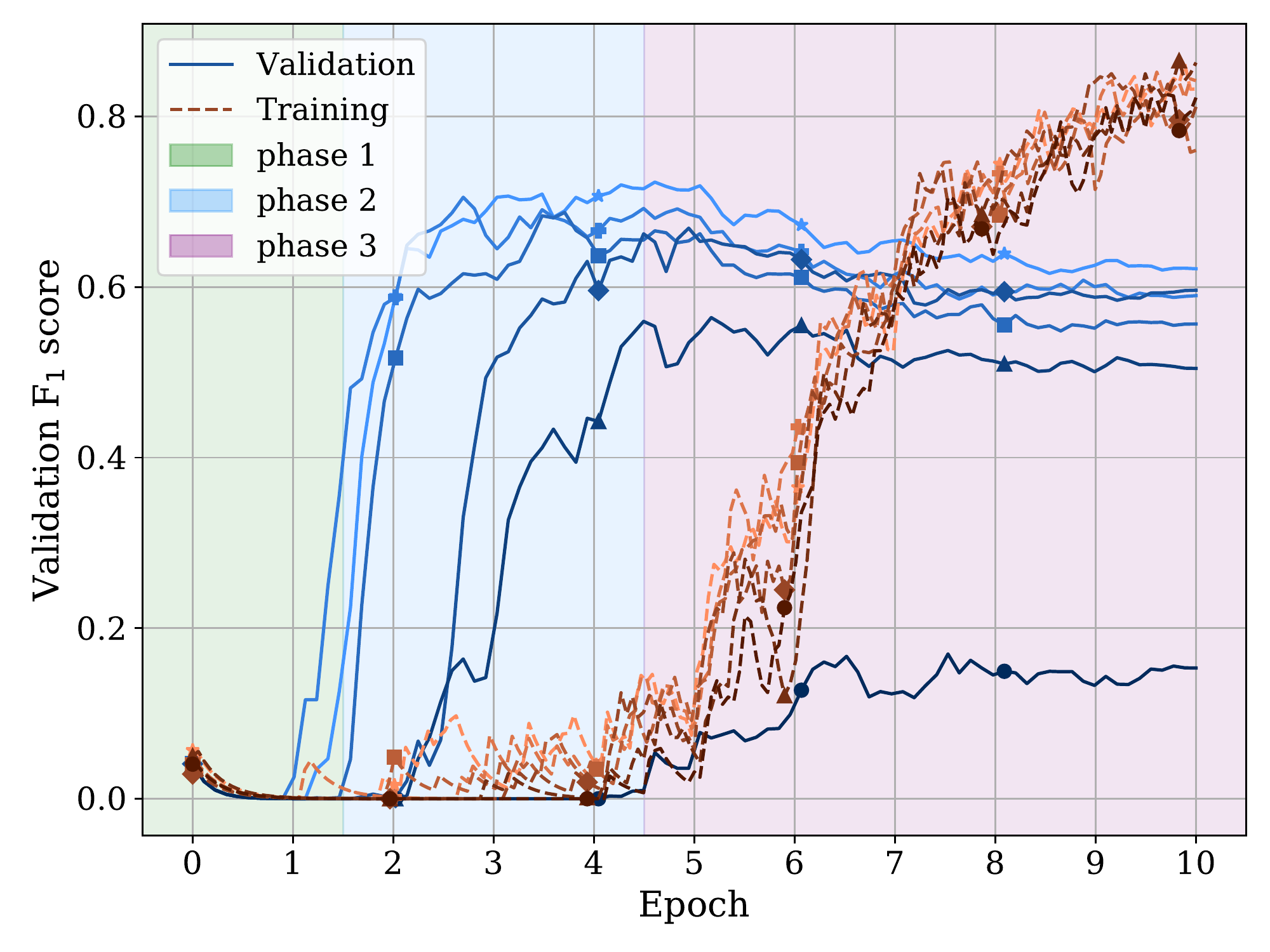}
    \caption{BERT performance (F$_1$) throughout the training process on the \texttt{CoNLL03-XMISC} train and validation sets. Darker colours correspond to fewer examples of the \texttt{MISC} class available (5 to 95 in steps of 20).}
    \label{fig:noisy-f1-misc-03}
\end{figure}

\begin{figure}[hbt]
	\centering
	\footnotesize
	\includegraphics[width=0.95\linewidth]{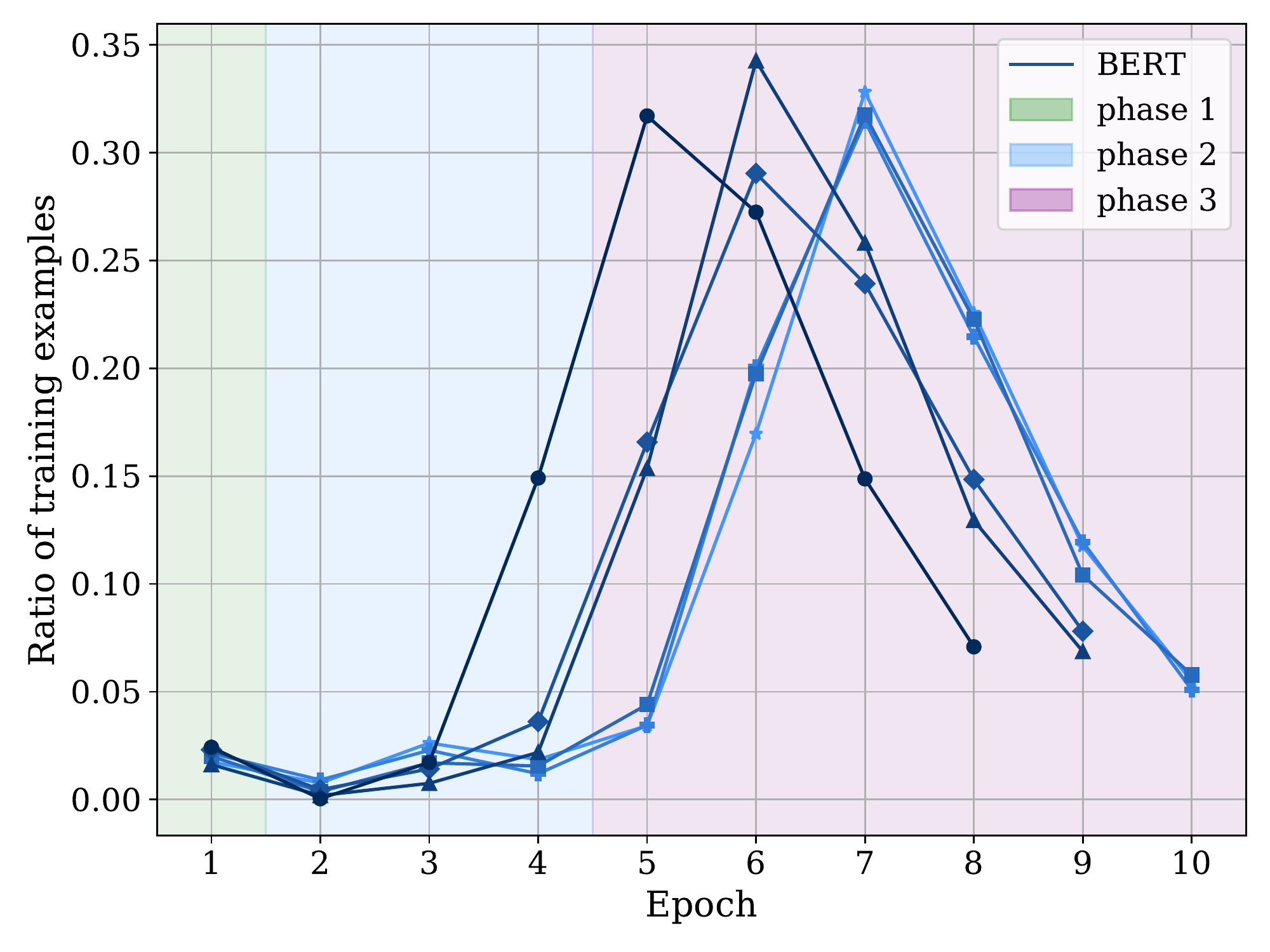}
    \caption{First learning events distribution during the training for various levels of noise on the \texttt{CoNLL03-XMISC} dataset. Darker colours correspond to fewer examples of the \texttt{MISC} class available (5 to 95 in steps of 20).}
    \label{fig:fles-misc-03}
\end{figure}

As per section \ref{sec:few-shot}, we also report the result of the experiments in the few-shot setting by removing most sentences containing the \texttt{MISC} class. The experimental setting is identical to the described in the main body of the paper. The relevant Figures are \ref{fig:noisy-f1-misc-03} and \ref{fig:fles-misc-03}.

\end{document}